\pdfoutput=1
\documentclass[11pt,dvipsnames]{article}

\usepackage[final]{main}

\usepackage{times}
\usepackage{latexsym}
\usepackage[whole]{bxcjkjatype} 
\usepackage{graphicx}
\usepackage{tabularx}
\usepackage{amsmath,amssymb}
\usepackage{mathtools}
\usepackage{here}
\usepackage{multirow}
\usepackage{booktabs,makecell,siunitx,eqparbox}
\usepackage[T1]{fontenc}
\usepackage[utf8]{inputenc}
\usepackage{inconsolata}
\usepackage{comment}
\usepackage{soul}
\usepackage{xcolor}
\usepackage{diagbox}
\usepackage{bm}
\usepackage{algorithmic}
\usepackage{algorithm}
\usepackage{array}
\usepackage{amssymb}
\usepackage[most]{tcolorbox}  
\usepackage{fancybox}
\usepackage{framed}
\tcbset{enhanced,colframe=red!50,colback=yellow!25,
nobeforeafter,shrink tight,left skip=1ex, right skip=1ex, extrude by=1.5mm}

\usepackage{booktabs}
\usepackage{cleveref}
\Crefname{section}{\S}{\S\S}
\crefname{table}{Table}{}
\crefname{figure}{Figure}{}
\crefname{algorithm}{Algorithm}{}
\crefname{equation}{eq.}{eqs.}
\crefname{appendix}{App.}{}
\crefname{prop}{Prop.}{}
\crefformat{section}{\S#2#1#3}  

\newcommand{\hlyg}[1]{{\sethlcolor{SpringGreen}\hl{#1}}}
\newcommand{\hlmelon}[1]{{\sethlcolor{Apricot}\hl{#1}}}
\newcolumntype{C}[1]{>{\centering\let\newline\\\arraybackslash\hspace{0pt}}m{#1}}

\title{First Heuristic Then Rational:  \\ Dynamic Use of Heuristics in Language Model Reasoning}
\author{Yoichi Aoki${}^{1,2}$,  
        Keito Kudo${}^{1,2}$,
        Tatsuki Kuribayashi${}^{3}$,\\ 
        {\bf Shusaku Sone${}^{1}$, Masaya Taniguchi${}^{2}$, Keisuke Sakaguchi${}^{1,2}$, Kentaro Inui${}^{3,1,2}$} \\
         ${}^{1}$Tohoku University,
         ${}^{2}$RIKEN, 
         ${}^{3}$MBZUAI \\ 
        \texttt{\{youichi.aoki.p2, keito.kudo.q4, sone.shusaku.r8\}@dc.tohoku.ac.jp, } \\
        \texttt{\{tatsuki.kuribayashi, kentaro.inui\}@mbzuai.ac.ae, } \\
        \texttt{keisuke.sakaguchi@tohoku.ac.jp,}
        \texttt{masaya.taniguchi@riken.jp} \\
        }

\begin{document}

\maketitle
%%%%%%%%%%%%%%%%%%%%%%%%%%%%%%%%%%%%%%
\begin{abstract}
\label{abstract}
Multi-step reasoning instruction, such as chain-of-thought prompting, is widely adopted to explore better language models (LMs) performance. We report on the systematic strategy that LMs employ in such a multi-step reasoning process.
Our controlled experiments reveal that LMs rely more heavily on heuristics, such as lexical overlap, in the earlier stages of reasoning, where more reasoning steps remain to reach a goal. 
Conversely, their reliance on heuristics decreases as LMs progress closer to the final answer through multiple reasoning steps. 
This suggests that LMs can backtrack only a limited number of future steps and dynamically combine heuristic strategies with rationale ones in tasks involving multi-step reasoning.\footnote{The code/data is available in \url{https://github.com/ao1neko/Heuristic-and-Rational-Reasoning}}
\end{abstract}
\section{Introduction}
When facing complex tasks, humans tend to seek shallow, heuristic solutions first~\cite{erickson1981semantic,frederick2005cognitive}. Once these attempts are revealed to fail or elicit another reasonable solution, they switch to being more rational~\cite{Stanovich_West_2000}.
Such systematic behavior helps us to predict how humans will tackle new problems.
Given such a view, when it comes to predicting the behavior of language models (LMs) (\citealt{Madaan2022textpattern,zhang2024how}; \textit{inter alia}), the following question naturally arises---Do LMs also use a similar \textit{systematic} strategy to solve complex tasks, or is their strategy totally different from humans, or do they have no such strategies? 
This study explores an answer to this question. 
Such analyses will shed light on the cognitive plausibility of LMs in problem solving~\cite{opedal2024do,eisape-etal-2024-systematic,Aher2022-uq} as well as address general concerns of current neural models relying on superficial, heuristic cues overly and ending up with irrational conclusions~\cite{du2022shortcut,lai2023whyshortcut,Palmer2017Adversarial,Ye2023lexical,Chen2024order}.

In this paper, we demonstrate that  LMs rely on shallow heuristics more frequently in the earlier phase of multi-step reasoning, and then gradually switch their reasoning strategy to be more rational and goal-oriented to make the right choice to reach the goal. 
From an engineering perspective, this highlights a limitation of modern LMs, including GPT-4~\cite{openai2023gpt4}, in searching for a solution at the initial stage of step-by-step reasoning, particularly when tasks require many-step-long solutions, implying that they can backtrack only a limited number of future steps from the answer to the current progress of reasoning.
From a cognitive perspective, their behaviors would be somewhat human-like in the sense that the models try to employ heuristics first in solving a complex problem.
Moreover, this paper is the first to show that models are equipped with both heuristic and goal-oriented reasoning and dynamically switch between them as needed.

\begin{figure}[t]
\centering
\includegraphics[width=\linewidth]{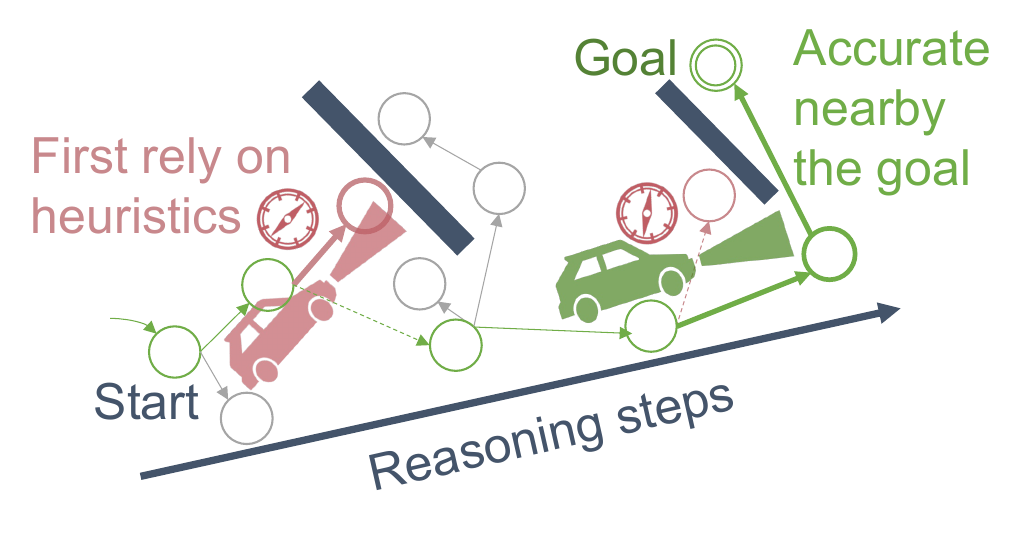}
\caption{Illustration of the systematic reasoning strategy we discovered in language models \includegraphics[width=1em]{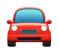}. When the goal is distant from the current reasoning step, they tend to rely on heuristics \includegraphics[width=0.8em]{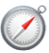} to take the next reasoning step, such as lexical overlap with a question, leading to the wrong direction (\textcolor{Salmon}{red path}). In contrast, when the goal is within a limited distance, they are more likely to take rational actions (\textcolor{LimeGreen}{green path}) to reach the goal.}
\label{fig:image0}
\end{figure}

%%%%%%%%%%%%%%%%%%%%%%%%%%%%%%%%%%%%%%%%%%%%%%%%
\begin{figure*}[t]
\centering
\includegraphics[width=\linewidth]{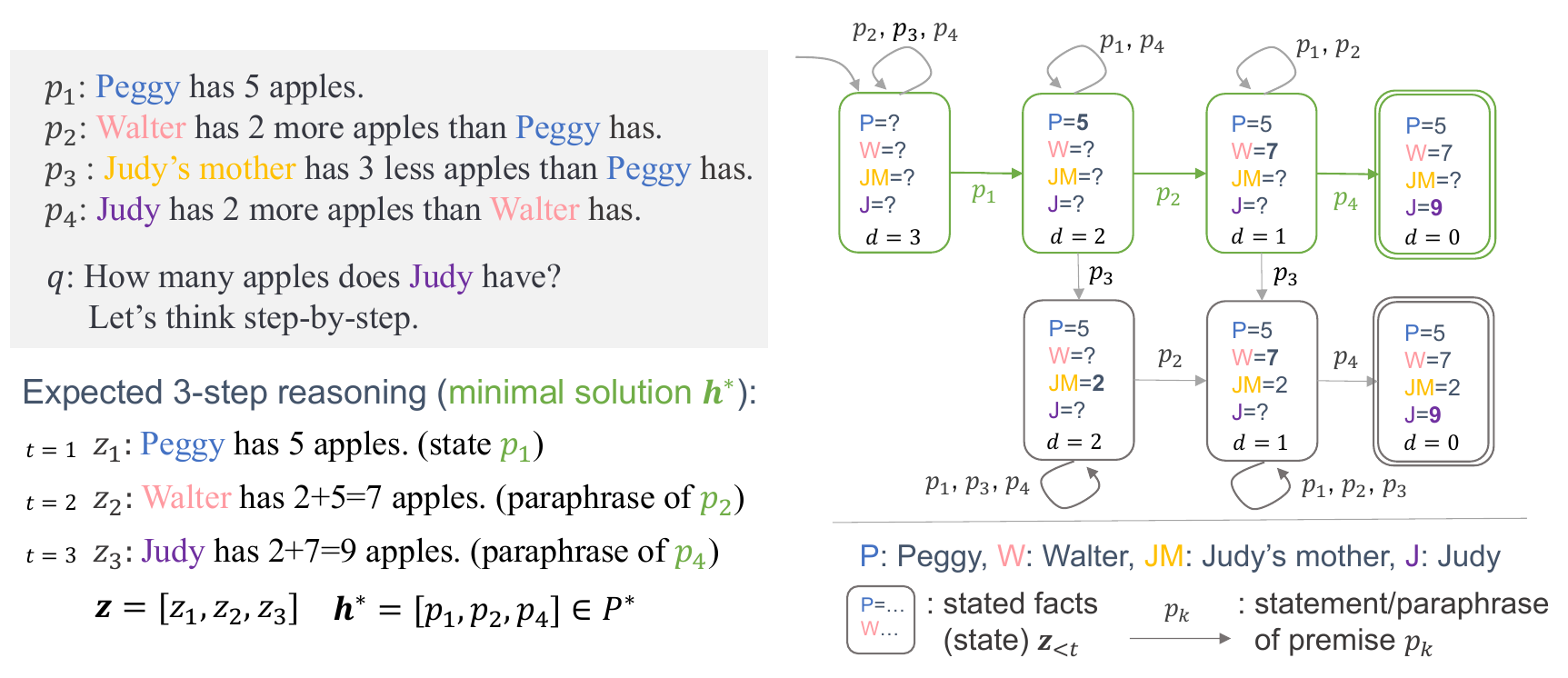}
\caption{Overview of the task setting. Given premises and a question, a model answers the question step-by-step (left part). Through each reasoning step $t$ of selecting/paraphrasing relevant premise $p_k \in P$, the available facts $\bm z$ are enriched (reasoning state progresses in the right part). If a reasoning step follows the minimal solution (\textcolor{LimeGreen}{green path} in the right part), the distance to the answer $d$ decreases.}
\label{fig:image1}
\end{figure*}
%%%%%%%%%%%%%%%%%%%%%%%%%%%
\section{Task}
\label{sec:definition}
We adopt an arithmetic reasoning task as a controlled testbed to analyze LM's reasoning ability (Figure~\ref{fig:image1} left).
We will use natural and artificially controlled datasets in the experiments, but let us use the latter, more formal examples to explain the task overview.

\paragraph{Arithmetic reasoning task:}
The problem consists of a set of premises $P=\{p_1,\cdots,p_k\}$ and a question $q$. 
Each premise describes either type of fact: (i) Person \texttt{A} has $n$ items (\texttt{A=$n$}), or (ii) Person \texttt{B} has $n$ more/fewer items than \texttt{A} has (\texttt{B=A+$n$} or \texttt{B=A-$n$}).   
The question $q$ asks the exact number of specific items a particular person ultimately has (\textit{How many items does B have?}).
Here, one should consider multiple premises to derive the final answer, e.g., \texttt{A=3; B=2+A; B=2+3=5}.
Notably, some premises are irrelevant to the answer; thus, models have to track which premise is necessary to reach the final answer.

\paragraph{Reasoning step:}
Let $f$ be a model that is instructed to solve the task step-by-step.
In each reasoning step $t$, the model $f$ selects a particular premise $p_i \in P$ and paraphrases it into a new fact $z_t$ by eagerly resolving reference expressions based on the already stated facts $\bm z_{<t}=[z_1,\cdots,z_{t-1}]$ as in equation (\ref{eq:step}):
\begin{equation}
    (p_i, z_t)=f(P,q,\bm z_{<t}) \;\;\,\mathrm{.}
    \label{eq:step}
\end{equation}
For example, in Figure~\ref{fig:image1}, when $p_2$, \textit{Walter has 2 more apples than Peggy.}, is selected at a particular reasoning step, the respective $z_t$ should be \textit{Walter has 2+5=7 apples.} if $\bm z_{<t}$ already contains the number of apples Peggy has, i.e., $p_1$.\footnote{If the reference can not be resolved with $\bm z_{<t}$, the model repeats the selected premise $p_i$ as $z_t$.} 
Starting with an empty set of stated facts $\bm z=\{\}$, the model recursively performs a reasoning step and can stop when outputting a special symbol \texttt{EOS} or answering the question $q$.
Here, we denote the whole history of selected premises as $\bm h=[p_i,\cdots,p_j] \in P^*$, where $P^{*}$ is Kleene closure of $P$.
Its $t$-th element $h_t$ is the premise to derive the $t$-th reasoning step $z_t$.
Henceforth, we call $\bm h$ \textit{reasoning steps} and focus on the ability to search for the right $\bm h$. 

\paragraph{Solutions:}
Among the possible reasoning steps $P^*$, there is a set of \textbf{solutions} $H^{\circ} \subset P^*$, where the final stated fact $z_{-1}$ in a solution $h \in H^{\circ}$ yields the right answer to the question $q$.
Figure~\ref{fig:image1} illustrates such a set of solutions $H^{\circ}$ as the steps 
 leading to the final states of the state transition graph  (right part of Figure~\ref{fig:image1}), e.g., $[p_1, p_3, p_2, p_4] \in H^{\circ}$.

\paragraph{Minimal solution:}
Within the set of solutions, there is only one \textbf{minimal solution} $\bm h^* \in H^{\circ} \subset P^*$.
Intuitively, $\bm h^*$ does not contain any irrelevant step to approach the answer; for example, the minimal solution of the problem in Figure~\ref{fig:image1} is $[p_1,p_2,p_4]=\bm h^*$.
To define  $\bm h^*$, let us first introduce a distance to the answer. In each reasoning step $t$, one can determine the minimum number of remaining reasoning steps to reach the answer $d \in \mathbb{N}$, given $\bm h_{\leq t}\in P^*$  and the initially provided premises $P$. 
The distance $d$ can be derived from a state transition graph and the minimum number of transitions to the closest final states, as shown in Figure~\ref{fig:image1} (right part).
Here, we denote the mapping function from $\bm h_{\leq t}$ to $d$ as $g: P^*\rightarrow \mathbb{N}$.
For example, $g([p_2,p_1,p_2])=1$ in Figure~\ref{fig:image1}.
A minimal solution $\bm h^*$ satisfies $\forall t\  g(\bm h^*_{\leq t})<g(\bm h^*_{\leq t-1})$.

\paragraph{Targeted ability of LMs:}
We evaluate LMs' ability to derive the minimal solution $\bm h^*$ as instructed by 4-shot examples (Table~\ref{table:fl_cot_prompt}).
Notably, we do not care about the ability to correctly introduce a new fact $z_t$ (Eq.~\ref{eq:step}), e.g., the accuracy of arithmetic operation (e.g., \texttt{5+2=7}), but separately focus on their search strategy to select the relevant premise to perform the next reasoning step.

\section{Heuristics}
\label{subsec:heuristics}
Given existing studies on LMs' use of heuristics (\cref{sec:related_work}), we focus on the following types of heuristics:

\paragraph{Lexical overlap between premise and question (\textsc{Overlap}):} Neural models generally tend to rely on superficial, shallow similarity of texts when considering their associations. We specifically examine whether models select premises with the same person name (PN) as the one in question as a representative of such biases.
For example, given a question \textit{how many apples \underline{Judy} has}, premises such as \textit{\underline{Judy}'s mother got 3 apples} might be selected as a relevant fact, regardless of its necessity to reach the answer. 

\paragraph{Position of premise (\textsc{Position}):}
It has been reported that models tend to select information, e.g., first and last, in specific positions in the context~\cite{10.1162/tacl_a_00638}.
We examine whether models tend to select the premise in the initial position of context.

\paragraph{Grammatical feature of premise (\textsc{Negative}):}
Given that a specific grammatical feature, e.g., negation word, is often a superficial cue~\cite{du2021nlumodel,niven2019probing}, we specifically analyze the bias that models avoid selecting premise with negation word, i.e., \textit{not}.
\section{Experiments}
\label{sec:experiment}
We first confirm that models indeed rely on particular types of heuristics in our setting (\cref{subsec:exp1}).
Then, we investigate \textit{when} in the step-by-step reasoning, such heuristics are more frequently exploited (\cref{subsec:exp2}). 

\paragraph{General settings:}
We use four representative variants of large language models (LLMs): text-bison-001 version of Google'sPaLM2~\cite{Anil2023palm2},  Llama2-13B~\cite{Touvron2023llama2}, gpt-3.5-turbo-0125 and gpt-4-0613 snapshots of OpenAI's GPT-3.5-turbo~\cite{openai2022gpt3} and GPT-4~\cite{openai2023gpt4}.
These models are instructed to yield a minimum solution via prompting.

\subsection{Preliminary experiments}
\label{subsec:exp1}
First, we confirm that LLMs exploit specific heuristics in natural and artificially controlled datasets during step-by-step reasoning.

\paragraph{Settings:}
We use two datasets as examples of multi-hop reasoning: GSM8K~\cite{cobbe2021gsm8k} (\cref{app:gsm8k}) and artificially-controlled dataset with 4-step arithmetic reasoning (\cref{app:exp1_data}).
To perform controlled experiments towards the LLMs' use of heuristics (\cref{sec:experiment}), we extend GSM8K and the artificially controlled dataset to the \textsc{Overlap}, \textsc{Possition}, and \textsc{Negative} variants. 
Each variant is created by adding one premise; there, the use of its corresponding heuristics make the model fail to find the minimal solution in step-by-step reasoning (\ref{eq:step}). For example, the \textsc{Position} dataset has an additional premise $\Tilde{p}$ (i.e., distractor) at the beginning of the sentences, which is irrelevant to answering the question $\Tilde{p} \not\in \bm h^*$.
The \textsc{Position} heuristics will lead the model to select the first sentence, i.e., distractor, as a neccesarry fact. 
% The added premise $\Tilde{p} \not\in \bm h^*$ is irrelevant to inducing the answer . 
As a baseline, we also created \textsc{Base} by adding a random distractor that does not match any of the three heuristics.
If the model more frequently selects the distractors in the \textsc{Overlap}, \textsc{Possition}, and \textsc{Negative} datasets, compared to \textsc{Base}, we can confirm that models are at least biased towards our selected heuristics.
We describe the details of the dataset creation process in Appendices~\cref{app:subsec:gsm8k_construction,app:subsec:exp1_data_construction}.
\begin{table}[t]
\centering
% \footnotesize
\fontsize{8}{10}\selectfont
\setlength{\tabcolsep}{0.7pt}
\begin{tabular}{lcccccccc}
\toprule
& \multicolumn{4}{c}{GSM8K} & \multicolumn{4}{c}{Artificial data} \\\cmidrule(r){2-5} \cmidrule(l){6-9}
Models     & Base   & Over. $\uparrow$ & Pos. $\uparrow$ & Neg. $\downarrow$ & Base   & Over. $\uparrow$ & Pos. $\uparrow$ & Neg. $\downarrow$  \\
     \cmidrule(r){1-1} \cmidrule(r){2-5} \cmidrule(l){6-9}
PaLM2   & 18.4\% & \textbf{57.9}\% & \textbf{19.7}\% & \textbf{17.1}\%  & 10.3\% & \textbf{42.3}\%   & \textbf{12.0}\%     & \textbf{4.3}\%   \\
Llama2 &43.2\% & \textbf{69.7}\% & \textbf{50.0}\% & \textbf{14.5}\% & 32.6\% & \textbf{67.7}\%   & \textbf{33.0}\%     & 41.7\%   \\
GPT-3.5 & 35.5\% & \textbf{67.1}\% & \textbf{36.8}\% & \textbf{22.4}\% & 21.0\% & 15.0\%   & \textbf{49.0}\%     & \textbf{0.0}\%   \\
GPT-4   & 21.1\% & \textbf{35.5}\% & \textbf{22.4}\% & 21.1\% & 0.0\%  & \textbf{0.01}\%   & 0.0\%      & 0.0\%   \\
\bottomrule
\end{tabular}
\caption{
   The frequency of the problems where the model selected a distractor $\Tilde{p}$ in step-by-step reasoning. ``Over.'', ``Pos.'' and ``Neg.'' denote the \textsc{Overlap}, \textsc{Position}, and \textsc{Negative} heuristics. The results are bolded when the frequency of output misled by the distractor increased (Over.↑ and Pos.↑) or decreased (Neg.↓) compared to the base results. In camera-ready, add the above explanation to the caption.
}
\label{table:FL_all_result}
\end{table}

\paragraph{Results:}
Table~\ref{table:FL_all_result} shows the frequencies of mentioning the distractor at least once during reasoning. 
The scores for \textsc{Overlap}, \textsc{Position}, and \textsc{Negative} in Table~\ref{table:FL_all_result})  are generally higher (lower for Negative) than the \textsc{Base} scores across models and datasets.
This indicates that LLMs, on average, tend to rely on our targeted heuristics (\cref{subsec:heuristics}).
Interestingly, different models yield different preferences towards distractor types; for example, Llama2 and GPT-3.5 have more biased premise positions than PaLM2 and GPT-4. 
Note that whether or not a distractor was selected was determined by some rules. Such details are exmplained in the Appendices~\cref{app:subsec:gsm8k_eval,app:subsec:exp1_data_eval}.

\subsection{Main experiments}
\label{subsec:exp2}
Then, we further investigate the LMs' dynamic use of heuristics.
We hypothesized that \textbf{the more distant the current reasoning step is from the answer (higher $d$ in~\cref{sec:definition}), the more heavily models rely on heuristics}.
% This is motivated by the assumption that a larger distance to the answer leads to more difficulty for the model in deriving the minimum solution (\cref{sec:definition}). 
Again, generating the minimal solution requires the model to track/plan the future path (remaining necessary and sufficient information) to reach the final answer, and its remaining length becomes longer at the initial phase of reasoning.
If models have a limited capacity to track the future path, they may have to give up rational reasoning and rely on heuristics, particularly at the earlier stage of reasoning, where the volume of the remaining future path is likely to exceed the model's capacity.
\begingroup
\begin{table}[t]
    \centering
    \small
    %\caption{実験7データセット：表層的な手がかり，論理的関係に基づいた推論に用いられやすい文によって構成される．（赤：表層的，黄：論理的）}
    \begin{tabular}{p{0.9\linewidth}}
        \toprule
        \textbf{Context:} \underline{Peggy} has 5 apples. \hlyg{Walter has 2 more apples than }\underline{\hlyg{Peggy}}. \hlmelon{Judy's mother has 3 less apples than }\underline{\hlmelon{Peggy.}} Judy has 4 more apples than Walter has. \\
        \textbf{Question:} How many apples does Judy have?\\
        \bottomrule
    \end{tabular}
        \caption{Example of a distractor examined in~\cref{subsec:exp2}. Suppose that $h^*_1$ is ``Peggy has 5 apples.'' Two candidate premises with ``Peggy'' seem to be plausible continuations as $h^*_2$, but only one is relevant to the final answer (green), and the other is a distractor (orange).}
    \label{table:step}
\end{table}
\endgroup

\begin{figure}[t]
\centering
\includegraphics[width=\linewidth]{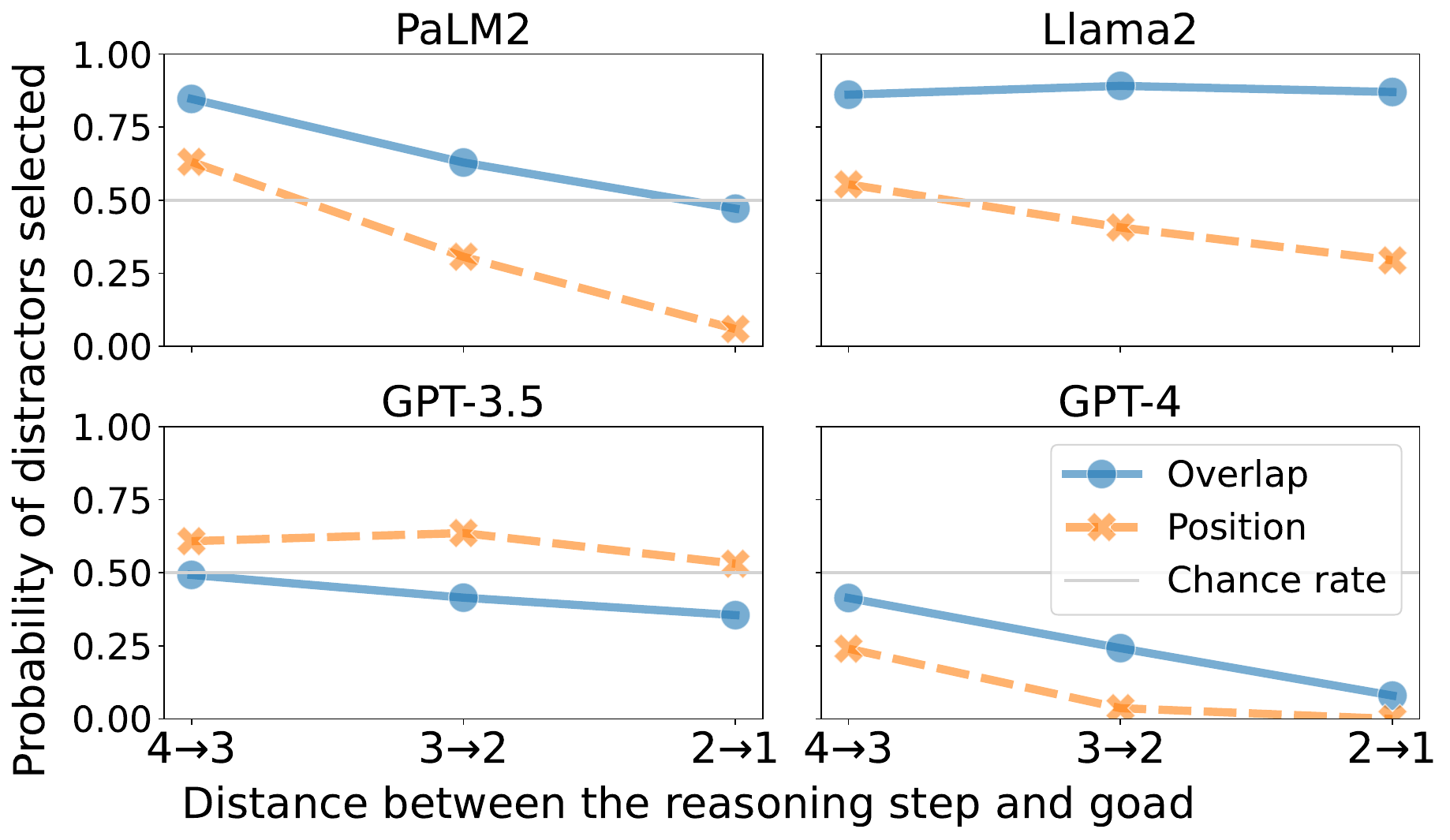}
\caption{The ratio at which a particular distractor is selected (y-axis: $r$) in each reasoning step (x-axis: $d$).}
\label{fig:step_results}
\end{figure}

\paragraph{Distractor and evaluation:}
To investigate the relationship between the distance to the answer and the models' reliance on heuristics, we identify at which steps heuristics are more likely to be exploited. Ideally, to facilitate a fair step-wise comparison, one should design a distractor \textit{equally} attractive to all the reasoning steps in $\bm h^*$ and analyze when it is selected during the reasoning; however, such a distractor is inherently difficult to implement.
Instead, we prepare multiple distractors $\tilde{P}$ to the problem in artificial data; each of them $\tilde{p}_t \in \tilde{P}$ correspond to each reasoning step $h^*_t$ in the sense that both share the same person name that appeared in the previous step $h^*_{t-1}$ (Table~\ref{table:step}).\footnote{To rule out the shortcut cue regarding the reference frequencies of each person name, we further added distractor premises to make the frequencies uniform.}
Similar to~\cref{subsec:exp1}, we further modify each distractor $\tilde{p}_t \in \tilde{P}$ to match each heuristic (Overlap with question or Position in~\cref{subsec:heuristics}).\footnote{We excluded the Negative (avoidance) bias because if a model avoids negation in the latter step, we cannot distinguish whether it was due to the heuristic or rational search.}
In evaluation, for each $t$, partial correct reasoning steps $\bm h^*_{<t}$ are teacher-forced to a model, and we analyze whether the model selects the right next step $h^*_t$ or its respective distractor $\tilde{p}_t$.
Specifically, we calculate the frequency $\#(\cdot)$ of models' selecting $\tilde{p}_t$  or  $h^*_t$; then, the ratio of exploiting a distractor $r=\frac{\#\tilde{p}_t}{\#\tilde{p}_t+\#h^*_t}$ is reported.
The chance rate should be 0.5. 

Note that we could not use the GSM8K dataset in this experiment since designing such controlled distractors for each reasoning step was not feasible, i.e., our creation of artificial, controlled data enables this kind of analysis.
In addition, we will not analyze the \textsc{Negative} heuristic in this experiment because it is a bias in the direction of avoidance, making the experimental design complicated.

\paragraph{Data:}
We used 5-step artificial reasoning data; that is, the distance to the answer is five at first $d=5$ and will monotonically decrease ($d=5\rightarrow 4\rightarrow \cdots \rightarrow0$) through the steps in the minimal solution. The first ($d=5\rightarrow4$) and the last ($d=1\rightarrow0$) steps are excluded from our analysis for their special properties; e.g., the right last step can be identified simply by detecting the lexical overlap with the question.

\paragraph{Results:}
The results are shown in Figure~\ref{fig:step_results}.
The x-axis is the change of remaining steps $d$ to the goal in the respective reasoning step, and the y-axis is the ratio $r$ of selecting distractors $\tilde{p}_t$.
The more distant the current step is from the answer (larger $d$), the more frequently the distractor is selected (larger $r$), which is typically above the chance rate. 
PaLM2 and GPT-4 exhibited apparent tendencies of the negative slopes between $d$ and $r$.
These results match the hypothesized behavior, the model's less rational behavior in earlier reasoning steps, and imply that they have a limited capacity to track the future reasoning path. 
\section{Related work}
\label{sec:related_work}
\paragraph{Multi-step symbolic reasoning:}
Given the general contrast between the symbolic and neural-based approaches in the artificial intelligence field~\cite{Hamilton2202neurosymbolic}, the community has questioned the ability of neural LMs in emulating particular symbolic operations, e.g., graph search algorithm~\cite{aoki-etal-2023-empirical,yao2023treeofthought,Fang2024neurosymbolic}.
In contrast, to identify what kind of symbolic tasks are (im)possible to solve for LMs by varying task complexities~\cite{Clark2020softreasoner}, we investigate the inherent, systematic biases in solving a particular symbolic reasoning task.

\paragraph{Heuristics in LM:}
Neural models have typically been distracted by superficial biases~\cite{du2022shortcut}. 
For example, they tend to use superficial linguistic artifacts~\cite{lai2023whyshortcut,sen2022whatmodel,du2021nlumodel,niven2019probing}, or more simply, positional features~\cite{ko2020firstsentence}, even with chain-of-thought prompting~\cite{Madaan2022textpattern}; these motivated our experiments. 
This paper shows that the LLMs' reliance on such heuristics changes dynamically as the reasoning progresses.

\paragraph{Search algorithm:}
Finding the shortest path between the start and the goal on a graph is a standard problem in computer science~\cite{russell2016artificial}.
Our investigation of LMs on the arithmetic tasks can be seen as characterizing LMs' biases as a search algorithm.
The use of heuristics in graph search is, more or less, related to the A* search algorithm~\cite{4082128}, although heuristics in A* search is a more narrow concept regarding the distance to the goal than those employed by LMs.

%%%%%%%%%%%%%%%%%%%%%%%%%%%%%%%%%%%%%%

\section{Conclusion}
We have found a systematic strategy for the use of heuristics in LMs' multi-step reasoning---a dynamic transition from a heuristic to a rational reasoning strategy during LMs' step-by-step reasoning. These results are hopefully helpful for researchers to understand their underlying mechanism as well as for LM users to consider the inherent biases systems have.

\section*{Limitations}
This study focused only on four specific language models and two arithmetic tasks.
Increasing the coverage of models and tasks is a possible future direction, although we ensured that our finding generalizes at least several models and task settings.
In~\cref{subsec:exp2}, we only used artificial datasets for designing controlled experiments to reduce confounding factors.
Constructing a controlled but natural dataset to evaluate the reasoning strategies of LMs should be encouraged.
Furthermore, our findings are based solely on the model's outputs, i.e., behaviors. Elucidating the underlying mechanisms inside the model and the source of these biases (e.g., statistical patterns in training data) should be investigated in future work.

\section*{Ethics statement}
This paper does not involve ethical concerns in the sense that we (i) did not conduct human experiments, (ii) just created artificial data without any potentially harmful contents, and (iii) did not address tasks related to ethically sensitive topics.

%%%%%%%%%%%%%%%%%%%%%%%%%%%%%%%%%%%%%%
\section*{Acknowledgements}
We thank the member of the Tohoku NLP Group for their cooperation in this research.
This work was supported by JSPS KAKENHI Grant Numbers 21K21343 and 24K16077, JST CREST Grant Number JPMJCR20D2, JST SPRING Grant Number JPMJSP2114, and JST BOOST Grant Number JPMJBS2421.

%%%%%%%%%%%%%%%%%%%%%%%%%%%%%%%%%%%%%%%%%%%%%%%%
\bibliography{custom}

\begin{thebibliography}{30}
\expandafter\ifx\csname natexlab\endcsname\relax\def\natexlab#1{#1}\fi

\bibitem[{Aher et~al.(2023)Aher, Arriaga, and Kalai}]{Aher2022-uq}
Gati~V. Aher, Rosa~I. Arriaga, and Adam~Tauman Kalai. 2023.
\newblock \href {https://proceedings.mlr.press/v202/aher23a.html} {Using large
  language models to simulate multiple humans and replicate human subject
  studies}.
\newblock In \emph{International Conference on Machine Learning, {ICML} 2023,
  23-29 July 2023, Honolulu, Hawaii, {USA}}, volume 202 of \emph{Proceedings of
  Machine Learning Research}, pages 337--371.

\bibitem[{Anil et~al.(2023)Anil, Dai, Firat, Johnson, Lepikhin, Passos,
  Shakeri, Taropa, Bailey, Chen, Chu, Clark, Shafey, Huang, Meier{-}Hellstern,
  Mishra, Moreira, Omernick, Robinson, Ruder, Tay, Xiao, Xu, Zhang,
  {\'{A}}brego, Ahn, Austin, Barham, Botha, Bradbury, Brahma, Brooks, Catasta,
  Cheng, Cherry, Choquette{-}Choo, Chowdhery, Crepy, Dave, Dehghani, Dev,
  Devlin, D{\'{\i}}az, Du, Dyer, Feinberg, Feng, Fienber, Freitag, Garcia,
  Gehrmann, Gonzalez, and et~al.}]{Anil2023palm2}
Rohan Anil, Andrew~M. Dai, Orhan Firat, Melvin Johnson, Dmitry Lepikhin,
  Alexandre Passos, Siamak Shakeri, Emanuel Taropa, Paige Bailey, Zhifeng Chen,
  Eric Chu, Jonathan~H. Clark, Laurent~El Shafey, Yanping Huang, Kathy
  Meier{-}Hellstern, Gaurav Mishra, Erica Moreira, Mark Omernick, Kevin
  Robinson, Sebastian Ruder, Yi~Tay, Kefan Xiao, Yuanzhong Xu, Yujing Zhang,
  Gustavo~Hern{\'{a}}ndez {\'{A}}brego, Junwhan Ahn, Jacob Austin, Paul Barham,
  Jan~A. Botha, James Bradbury, Siddhartha Brahma, Kevin Brooks, Michele
  Catasta, Yong Cheng, Colin Cherry, Christopher~A. Choquette{-}Choo, Aakanksha
  Chowdhery, Cl{\'{e}}ment Crepy, Shachi Dave, Mostafa Dehghani, Sunipa Dev,
  Jacob Devlin, Mark D{\'{\i}}az, Nan Du, Ethan Dyer, Vladimir Feinberg,
  Fangxiaoyu Feng, Vlad Fienber, Markus Freitag, Xavier Garcia, Sebastian
  Gehrmann, Lucas Gonzalez, and et~al. 2023.
\newblock \href {https://doi.org/10.48550/ARXIV.2305.10403} {Palm 2 technical
  report}.
\newblock \emph{CoRR}, abs/2305.10403.

\bibitem[{Aoki et~al.(2023)Aoki, Kudo, Kuribayashi, Brassard, Yoshikawa,
  Sakaguchi, and Inui}]{aoki-etal-2023-empirical}
Yoichi Aoki, Keito Kudo, Tatsuki Kuribayashi, Ana Brassard, Masashi Yoshikawa,
  Keisuke Sakaguchi, and Kentaro Inui. 2023.
\newblock \href {https://doi.org/10.18653/v1/2023.findings-eacl.86} {Empirical
  investigation of neural symbolic reasoning strategies}.
\newblock In \emph{Findings of the European Chapter of the Association for
  Computational Linguistics: EACL 2023}, pages 1154--1162, Dubrovnik, Croatia.

\bibitem[{Chen et~al.(2024)Chen, Chi, Wang, and Zhou}]{Chen2024order}
Xinyun Chen, Ryan~A. Chi, Xuezhi Wang, and Denny Zhou. 2024.
\newblock \href {https://doi.org/10.48550/ARXIV.2402.08939} {Premise order
  matters in reasoning with large language models}.
\newblock \emph{CoRR}, abs/2402.08939.

\bibitem[{Clark et~al.(2020)Clark, Tafjord, and
  Richardson}]{Clark2020softreasoner}
Peter Clark, Oyvind Tafjord, and Kyle Richardson. 2020.
\newblock \href {https://doi.org/10.24963/IJCAI.2020/537} {Transformers as soft
  reasoners over language}.
\newblock In \emph{Proceedings of the Twenty-Ninth International Joint
  Conference on Artificial Intelligence, {IJCAI} 2020}, pages 3882--3890.

\bibitem[{Cobbe et~al.(2021)Cobbe, Kosaraju, Bavarian, Chen, Jun, Kaiser,
  Plappert, Tworek, Hilton, Nakano, Hesse, and Schulman}]{cobbe2021gsm8k}
Karl Cobbe, Vineet Kosaraju, Mohammad Bavarian, Mark Chen, Heewoo Jun, Lukasz
  Kaiser, Matthias Plappert, Jerry Tworek, Jacob Hilton, Reiichiro Nakano,
  Christopher Hesse, and John Schulman. 2021.
\newblock \href {http://arxiv.org/abs/2110.14168} {Training verifiers to solve
  math word problems}.
\newblock \emph{CoRR}, abs/2110.14168.

\bibitem[{Du et~al.(2022)Du, He, Zou, Tao, and Hu}]{du2022shortcut}
Mengnan Du, Fengxiang He, Na~Zou, Dacheng Tao, and Xia Hu. 2022.
\newblock \href {https://doi.org/10.48550/ARXIV.2208.11857} {Shortcut learning
  of large language models in natural language understanding: {A} survey}.
\newblock \emph{CoRR}, abs/2208.11857.

\bibitem[{Du et~al.(2021)Du, Manjunatha, Jain, Deshpande, Dernoncourt, Gu, Sun,
  and Hu}]{du2021nlumodel}
Mengnan Du, Varun Manjunatha, Rajiv Jain, Ruchi Deshpande, Franck Dernoncourt,
  Jiuxiang Gu, Tong Sun, and Xia Hu. 2021.
\newblock \href {https://doi.org/10.18653/V1/2021.NAACL-MAIN.71} {Towards
  interpreting and mitigating shortcut learning behavior of {NLU} models}.
\newblock In \emph{Proceedings of the 2021 Conference of the North American
  Chapter of the Association for Computational Linguistics: Human Language
  Technologies, {NAACL-HLT} 2021}, pages 915--929.

\bibitem[{Eisape et~al.(2024)Eisape, Tessler, Dasgupta, Sha, Steenkiste, and
  Linzen}]{eisape-etal-2024-systematic}
Tiwalayo Eisape, Michael Tessler, Ishita Dasgupta, Fei Sha, Sjoerd Steenkiste,
  and Tal Linzen. 2024.
\newblock \href {https://doi.org/10.18653/v1/2024.naacl-long.466} {A systematic
  comparison of syllogistic reasoning in humans and language models}.
\newblock In \emph{Proceedings of the 2024 Conference of the North American
  Chapter of the Association for Computational Linguistics: Human Language
  Technologies, NAACL-HLT 2024}, pages 8425--8444.

\bibitem[{Erickson and Mattson(1981)}]{erickson1981semantic}
Thomas~D. Erickson and Mark~E. Mattson. 1981.
\newblock \href
  {https://www.sciencedirect.com/science/article/pii/S0022537181901651?via%3Dihub}
  {From words to meaning: A semantic illusion}.
\newblock \emph{Journal of Verbal Learning and Verbal Behavior},
  20(5):540--551.

\bibitem[{Fang et~al.(2024)Fang, Deng, Zhang, Shi, Chen, Pechenizkiy, and
  Wang}]{Fang2024neurosymbolic}
Meng Fang, Shilong Deng, Yudi Zhang, Zijing Shi, Ling Chen, Mykola Pechenizkiy,
  and Jun Wang. 2024.
\newblock \href {https://doi.org/10.1609/AAAI.V38I16.29754} {Large language
  models are neurosymbolic reasoners}.
\newblock In \emph{Proceedings of the AAAI Conference on Artificial
  Intelligence}, pages 17985--17993.

\bibitem[{Frederick(2005)}]{frederick2005cognitive}
Shane Frederick. 2005.
\newblock \href {https://www.aeaweb.org/articles?id=10.1257/089533005775196732}
  {Cognitive reflection and decision making}.
\newblock \emph{Journal of Economic Perspectives}, 19(4):25--42.

\bibitem[{Hamilton et~al.(2022)Hamilton, Nayak, Bozic, and
  Longo}]{Hamilton2202neurosymbolic}
Kyle Hamilton, Aparna Nayak, Bojan Bozic, and Luca Longo. 2022.
\newblock \href {http://arxiv.org/abs/2202.12205} {Is neuro-symbolic {AI}
  meeting its promise in natural language processing? {A} structured review}.
\newblock \emph{CoRR}, abs/2202.12205.

\bibitem[{Hart et~al.(1968)Hart, Nilsson, and Raphael}]{4082128}
Peter~E. Hart, Nils~J. Nilsson, and Bertram Raphael. 1968.
\newblock \href {https://doi.org/10.1109/TSSC.1968.300136} {A formal basis for
  the heuristic determination of minimum cost paths}.
\newblock \emph{IEEE Transactions on Systems Science and Cybernetics},
  4(2):100--107.

\bibitem[{Jia and Liang(2017)}]{Palmer2017Adversarial}
Robin Jia and Percy Liang. 2017.
\newblock \href {https://doi.org/10.18653/V1/D17-1215} {Adversarial examples
  for evaluating reading comprehension systems}.
\newblock In \emph{Proceedings of the 2017 Conference on Empirical Methods in
  Natural Language Processing, {EMNLP} 2017}, pages 2021--2031.

\bibitem[{Ko et~al.(2020)Ko, Lee, Kim, Kim, and Kang}]{ko2020firstsentence}
Miyoung Ko, Jinhyuk Lee, Hyunjae Kim, Gangwoo Kim, and Jaewoo Kang. 2020.
\newblock \href {https://doi.org/10.18653/V1/2020.EMNLP-MAIN.84} {Look at the
  first sentence: Position bias in question answering}.
\newblock In \emph{Proceedings of the 2020 Conference on Empirical Methods in
  Natural Language Processing, {EMNLP} 2020}, pages 1109--1121.

\bibitem[{Lai et~al.(2021)Lai, Zhang, Feng, Huang, and
  Zhao}]{lai2023whyshortcut}
Yuxuan Lai, Chen Zhang, Yansong Feng, Quzhe Huang, and Dongyan Zhao. 2021.
\newblock \href {https://doi.org/10.18653/V1/2021.FINDINGS-ACL.85} {Why machine
  reading comprehension models learn shortcuts?}
\newblock In \emph{Findings of the Association for Computational Linguistics:
  {ACL/IJCNLP} 2021}, pages 989--1002.

\bibitem[{Liu et~al.(2024)Liu, Lin, Hewitt, Paranjape, Bevilacqua, Petroni, and
  Liang}]{10.1162/tacl_a_00638}
Nelson~F. Liu, Kevin Lin, John Hewitt, Ashwin Paranjape, Michele Bevilacqua,
  Fabio Petroni, and Percy Liang. 2024.
\newblock \href {https://doi.org/10.1162/tacl_a_00638} {{Lost in the Middle:
  How Language Models Use Long Contexts}}.
\newblock \emph{Transactions of the Association for Computational Linguistics},
  12:157--173.

\bibitem[{Madaan and Yazdanbakhsh(2022)}]{Madaan2022textpattern}
Aman Madaan and Amir Yazdanbakhsh. 2022.
\newblock \href {https://doi.org/10.48550/ARXIV.2209.07686} {Text and patterns:
  For effective chain of thought, it takes two to tango}.
\newblock \emph{CoRR}, abs/2209.07686.

\bibitem[{Niven and Kao(2019)}]{niven2019probing}
Timothy Niven and Hung{-}Yu Kao. 2019.
\newblock \href {https://doi.org/10.18653/V1/P19-1459} {Probing neural network
  comprehension of natural language arguments}.
\newblock In \emph{Proceedings of the 57th Conference of the Association for
  Computational Linguistics, {ACL} 2019, Volume 1}, pages 4658--4664.
  Association for Computational Linguistics.

\bibitem[{Opedal et~al.(2024)Opedal, Stolfo, Shirakami, Jiao, Cotterell,
  Sch{\"o}lkopf, Saparov, and Sachan}]{opedal2024do}
Andreas Opedal, Alessandro Stolfo, Haruki Shirakami, Ying Jiao, Ryan Cotterell,
  Bernhard Sch{\"o}lkopf, Abulhair Saparov, and Mrinmaya Sachan. 2024.
\newblock \href {https://openreview.net/forum?id=k1JXxbpIY6} {Do language
  models exhibit the same cognitive biases in problem solving as human
  learners?}
\newblock In \emph{Forty-first International Conference on Machine Learning}.

\bibitem[{OpenAI(2022)}]{openai2022gpt3}
OpenAI. 2022.
\newblock \href {https://openai.com/blog/chatgpt} {Introducing chatgpt.}

\bibitem[{OpenAI(2023)}]{openai2023gpt4}
OpenAI. 2023.
\newblock \href {https://doi.org/10.48550/ARXIV.2303.08774} {{GPT-4} technical
  report}.
\newblock \emph{CoRR}, abs/2303.08774.

\bibitem[{Russell and Norvig(1995)}]{russell2016artificial}
Stuart~J. Russell and Peter Norvig. 1995.
\newblock \href {https://www.worldcat.org/oclc/31288015} {\emph{Artificial
  intelligence - a modern approach: the intelligent agent book}}.
\newblock Prentice Hall series in artificial intelligence. Prentice Hall.

\bibitem[{Sen and Saffari(2020)}]{sen2022whatmodel}
Priyanka Sen and Amir Saffari. 2020.
\newblock \href {https://doi.org/10.18653/V1/2020.EMNLP-MAIN.190} {What do
  models learn from question answering datasets?}
\newblock In \emph{Proceedings of the 2020 Conference on Empirical Methods in
  Natural Language Processing, {EMNLP} 2020}, pages 2429--2438.

\bibitem[{Stanovich and West(2000)}]{Stanovich_West_2000}
Keith~E. Stanovich and Richard~F. West. 2000.
\newblock \href {https://doi.org/10.1017/S0140525X00003435} {Individual
  differences in reasoning: Implications for the rationality debate?}
\newblock \emph{Behavioral and Brain Sciences}, 23(5):645^^e2^^80^^93665.

\bibitem[{Touvron et~al.(2023)Touvron, Martin, Stone, Albert, Almahairi,
  Babaei, Bashlykov, Batra, Bhargava, Bhosale, Bikel, Blecher, Canton{-}Ferrer,
  Chen, Cucurull, Esiobu, Fernandes, Fu, Fu, Fuller, Gao, Goswami, Goyal,
  Hartshorn, Hosseini, Hou, Inan, Kardas, Kerkez, Khabsa, Kloumann, Korenev,
  Koura, Lachaux, Lavril, Lee, Liskovich, Lu, Mao, Martinet, Mihaylov, Mishra,
  Molybog, Nie, Poulton, Reizenstein, Rungta, Saladi, Schelten, Silva, Smith,
  Subramanian, Tan, Tang, Taylor, Williams, Kuan, Xu, Yan, Zarov, Zhang, Fan,
  Kambadur, Narang, Rodriguez, Stojnic, Edunov, and
  Scialom}]{Touvron2023llama2}
Hugo Touvron, Louis Martin, Kevin Stone, Peter Albert, Amjad Almahairi, Yasmine
  Babaei, Nikolay Bashlykov, Soumya Batra, Prajjwal Bhargava, Shruti Bhosale,
  Dan Bikel, Lukas Blecher, Cristian Canton{-}Ferrer, Moya Chen, Guillem
  Cucurull, David Esiobu, Jude Fernandes, Jeremy Fu, Wenyin Fu, Brian Fuller,
  Cynthia Gao, Vedanuj Goswami, Naman Goyal, Anthony Hartshorn, Saghar
  Hosseini, Rui Hou, Hakan Inan, Marcin Kardas, Viktor Kerkez, Madian Khabsa,
  Isabel Kloumann, Artem Korenev, Punit~Singh Koura, Marie{-}Anne Lachaux,
  Thibaut Lavril, Jenya Lee, Diana Liskovich, Yinghai Lu, Yuning Mao, Xavier
  Martinet, Todor Mihaylov, Pushkar Mishra, Igor Molybog, Yixin Nie, Andrew
  Poulton, Jeremy Reizenstein, Rashi Rungta, Kalyan Saladi, Alan Schelten, Ruan
  Silva, Eric~Michael Smith, Ranjan Subramanian, Xiaoqing~Ellen Tan, Binh Tang,
  Ross Taylor, Adina Williams, Jian~Xiang Kuan, Puxin Xu, Zheng Yan, Iliyan
  Zarov, Yuchen Zhang, Angela Fan, Melanie Kambadur, Sharan Narang,
  Aur{\'{e}}lien Rodriguez, Robert Stojnic, Sergey Edunov, and Thomas Scialom.
  2023.
\newblock \href {https://doi.org/10.48550/ARXIV.2307.09288} {Llama 2: Open
  foundation and fine-tuned chat models}.
\newblock \emph{CoRR}, abs/2307.09288.

\bibitem[{Yao et~al.(2023)Yao, Yu, Zhao, Shafran, Griffiths, Cao, and
  Narasimhan}]{yao2023treeofthought}
Shunyu Yao, Dian Yu, Jeffrey Zhao, Izhak Shafran, Thomas~L. Griffiths, Yuan
  Cao, and Karthik Narasimhan. 2023.
\newblock \href {https://doi.org/10.48550/ARXIV.2305.10601} {Tree of thoughts:
  Deliberate problem solving with large language models}.
\newblock \emph{CoRR}, abs/2305.10601.

\bibitem[{Ye et~al.(2023)Ye, Kuribayashi, Suzuki, Kobayashi, and
  Funayama}]{Ye2023lexical}
Mengyu Ye, Tatsuki Kuribayashi, Jun Suzuki, Goro Kobayashi, and Hiroaki
  Funayama. 2023.
\newblock \href {https://doi.org/10.18653/V1/2023.EMNLP-MAIN.912} {Assessing
  step-by-step reasoning against lexical negation: {A} case study on
  syllogism}.
\newblock In \emph{Proceedings of the 2023 Conference on Empirical Methods in
  Natural Language Processing, {EMNLP} 2023}, pages 14753--14773.

\bibitem[{Zhang et~al.(2024)Zhang, Press, Merrill, Liu, and
  Smith}]{zhang2024how}
Muru Zhang, Ofir Press, William Merrill, Alisa Liu, and Noah~A. Smith. 2024.
\newblock \href {https://openreview.net/forum?id=FPlaQyAGHu} {How language
  model hallucinations can snowball}.
\newblock In \emph{Forty-first International Conference on Machine Learning}.

\end{thebibliography}
\bibliographystyle{acl_natbib}
%%%%%%%%%%%%%%%%%%%%%%%%%%%%%%%%%%%%%%%%%%%%%%%%
\clearpage
\appendix
\section{GSM8K experiments in~\cref{subsec:exp1}}
\label{app:gsm8k}

\subsection{Dataset construction process}
\label{app:subsec:gsm8k_construction}
As described in~\ref{subsec:exp1}, we modify the existing multi-hop numerical reasoning dataset, GSM8K (distributed under the MIT license), to construct the evaluation dataset.
The dataset construction process is divided into two steps: 1. Extracting instances from GSM8K, and 2. Inserting distractors according to the heuristic we want to evaluate for each extracted problem statement.

\subsubsection{Instance extraction}
\label{app:subsubsec:sample_extraction}
\begingroup
\begin{table}[t]
    \centering
    \small
    \begin{tabular}{p{0.9\linewidth}}
        \toprule
        \textbf{Context:} \underline{James}$_{name}$ decides to run \underline{3}$_{num}$ sprints \underline{3}$_{num}$ times a week. \underline{He}$_{pronoun}$ runs \underline{60}$_{num}$ meters each sprint. \\
        \\
        \textbf{Question:} How many total meters does \underline{he}$_{pronoun}$ run a week?\\
        \midrule
        \textbf{Person's Names：} James\\
        \textbf{Numbers：} 3,60 \\
        \bottomrule
    \end{tabular}
        \caption{
     Extraction of names, personal pronouns, and numbers on GSM8K.
    }
    \label{table:appendix_GSM8K}
\end{table}
\endgroup

There is no guarantee that the premises added during data expansion will not affect the solution. Therefore, instances are extracted in which the addition of premises has little effect on the solution. Specifically, we extract instances from the GSM8K evaluation dataset following the process below:
\begin{enumerate}
\item We manually create a list of 50 person names (PNs) from a subset of the GSM8K evaluation dataset.
\item Using regular expressions, we identify PNs from this list.
\item We identify pronouns and numerical expressions present in each instance.
\item We extracted instances that included precisely one from our list in both the context and the question, and where either the PN or a pronoun appears in the question (e.g., Table~\ref{table:appendix_GSM8K}).
\item We replaced all pronouns within the extracted instances with the corresponding person’s name.
\end{enumerate}
In the following process, information on persons not appearing in the problem statement is added as distractors. The instance questions extracted above are about persons in the problem statement. Therefore, it can be guaranteed that adding a distractor will not change the answer by such a process.

\subsubsection{Distractor insertion}
Subsequently, we added distractors to the extracted instances according to each heuristic, thereby constructing 76 instances for the evaluation dataset.
Below, we will describe the process of creating the evaluation dataset for each heuristic.

\paragraph{Base}
We insert a template-based random distractor (i.e., $\Tilde{p}$) into each instance as a baseline.
The distractor was created using the following steps:
\begin{enumerate}
\item We randomly selected one sentence from the instance that included a PN or pronoun and a numerical expression.
\item We replaced the PN or pronoun in the selected sentence with a placeholder, \lbrack name\rbrack.
\item We replaced the numerical expression in the selected sentence with a placeholder, \lbrack num\rbrack.
\item We replaced \lbrack name\rbrack \, with a randomly selected name from the list of PNs created in~\ref{app:subsubsec:sample_extraction}, excluding the name already present in the instance.
\item We replaced \lbrack num\rbrack \, with another value.\footnote{The replacement number was calculated by multiplying each number appearing in the sentence by either 0.5, 0.8, 1.2, 1.5, or 2 and then rounding down to the nearest whole number.}
\item We inserted the created distractor into a random position in the context other than the beginning of the instance.
\end{enumerate}
For example, When the sentence ``James decides to run 3 sprints 3 times a week.'' is selected from the instance in Table~\ref{table:appendix_GSM8K}, a template ``\lbrack name\rbrack \, decides to run \lbrack num\rbrack \, sprints \lbrack num\rbrack \, times a week.'' is crafted. Names and numbers are randomly selected from the candidates and placed into these placeholders, and the resulting distractor is then inserted into the context.

\paragraph{Overlap}
To evaluate whether the Overlap heuristic influences the model, we insert distractors $\Tilde{p}$ into each instance following the steps below:
\begin{enumerate}
\item We substituted the placeholder \lbrack name\rbrack \, within the Base distractor template with the person’s name found in the instance, appended by relational phrases such as ``'s mother'', ``'s father'', ``'s son'', or ``'s neighborhood'' (e.g., in the instance from Table~\ref{table:appendix_GSM8K}, this would become ``James's mother'').
\item We replaced the number in the sentence with another numerical value.
\item We placed the constructed distractor into context at the location where the Base distractor was positioned in the instance.
\end{enumerate}

\paragraph{Position}
To evaluate whether the heuristic of Position induces the model, we insert distractors $\Tilde{p}$ into each instance.
Each distractor is identical to the Base distractor except for its insertion point.
Specifically, we relocated the distractor's insertion point to a random position closer to the beginning of the context than the position used for the Base distractor.

\paragraph{Negative}
To evaluate the model's response to the Negative heuristic, we insert distractors $\Tilde{p}$ into each instance created based on the following template:

\begin{center}
  \lbrack name\rbrack \, doesn't have [num] [object].\\
\end{center}

In this template:
\begin{itemize}
\item \lbrack name\rbrack \, is substituted with a random PN included in the instance.
\item \lbrack object\rbrack \, is replaced with one of the following items: ``apples,'' ``bananas,'' ``grapes,'' ``pencils,'' or ``books.''
\item \lbrack num\rbrack \, is replaced with a different numerical value, using the same algorithm for creating the Base distractor.
\end{itemize}

\subsection{Evaluation}
\label{app:subsec:gsm8k_eval}
To determine whether the LMs selected the distractor during reasoning, we check if the numbers in the distractor $\Tilde{p}$ are in the facts $\bm{z}$.
We calculate the frequency of instances where the distractor is selected.

\section{Artificial-data experiments in~\cref{subsec:exp1}}
\label{app:exp1_data}

\subsection{Dataset construction process}
\label{app:subsec:exp1_data_construction}
\paragraph{Base}
We construct the artificial data using the method outlined below, based on the template presented in Table~\ref{table:appendix_FL1}.
\begin{itemize}
\item Randomly assign one of the following names to the placeholders \lbrack nameA\rbrack \, to \lbrack nameE\rbrack : ``Alice,'' ``Bob,'' ``Carol,'' ``Dave,'' ``Eve,'' ``Frank,'' ``Grace,'' ``Heidi,'' ``Ivan,'' ``Judy,'' ``Kevin,'' ``Larry,'' ``Mallory,'' ``Nancy,'' ``Olivia,'' ``Peggy,'' ``Quentin,'' ``Rob,'' ``Sybil,'' ``Trent,'' ``Ursula,'' ``Victor,'' ``Walter,'' ``Xavier,'' ``Yvonne,'' or ``Zoe.''
\item Assign a randomly selected value from \lbrack nameA\rbrack \, to \lbrack nameD\rbrack \, to the placeholder \lbrack nameX\rbrack.
\item Assign a random number from 0 to 100 to the placeholder \lbrack num\rbrack.
\item Assign one of the objects ``apples,'' ``bananas,'' ``grapes,'' ``pencils,'' or ``books'' to the placeholder \lbrack object\rbrack.
\item Assign either ``more'' or ``fewer'' to the placeholder \lbrack relation\rbrack.
\item Randomly shuffle the order of the sentences.
\end{itemize}
\begingroup
\begin{table}[t]
    \centering
    \small
    \begin{tabular}{p{0.9\linewidth}}
        \toprule
Context：[nameA] has [num] [object].\\
\lbrack nameB] has [num] [relation] [object] than [nameA].\\
\lbrack nameC] has [num] [relation] [object] than [nameB].\\
\lbrack nameD] has [num] [relation] [object] than [nameC].\\\\
Question：How many [object] does [nameD] have?\\\\
distractor：[nameE] has [num] [relation] [object] than [nameX].\\
        \bottomrule
    \end{tabular}
        \caption{
    Template of artificial data in~\cref{subsec:exp1}.
    }
    \label{table:appendix_FL1}
\end{table}
\endgroup

\paragraph{Overlap}
We constructed a dataset to evaluate whether the Overlap heuristic induced the model by making certain modifications to the Base distractor for each instance.
Specifically, we modified the value of \lbrack nameD\rbrack \, by appending relational phrases such as `` 's mother'', `` 's father'', `` 's son'', or `` 's neighborhood'' to the existing value of \lbrack nameD\rbrack.
We then assigned this modified value to \lbrack nameE\rbrack.

\paragraph{Position}
We modify the Base distractor to evaluate if the Position heuristic induces the model.
Specifically, we altered the insertion point of the Base distractor to a randomly chosen position closer to the context's beginning than the original position used in the Base distractor.

\paragraph{Negative}
To evaluate whether the Neg induces the model heuristic, we construct a dataset by modifying the Base distractor.
Specifically, we convert the Base distractor into a negative expression (e.g., \lbrack nameE\rbrack \, \emph{doesn't have} \lbrack num\rbrack \, \lbrack relation\rbrack \, \lbrack object\rbrack \, than \lbrack nameX\rbrack).

\subsection{Evaluation}
\label{app:subsec:exp1_data_eval}
To determine whether the LMs selected the distractor during reasoning, we check if the subject of the distractor (i.e., \lbrack nameE\rbrack) is included in the facts $\bm{z}$.
We calculate the frequency of instances where the distractor is selected.

\section{Artificial data in ~\cref{subsec:exp2}}
\label{app:exp2_data}
\begingroup
\begin{table}[t]
    \centering
    \small
    \begin{tabular}{p{0.9\linewidth}}
        \toprule
Cntext：[nameA] has [num] [object].\\
\lbrack nameB] has [num] [relation] [object] than [nameA].\\
\lbrack nameC] has [num] [relation] [object] than [nameB].\\
\lbrack nameD] has [num] [relation] [object] than [nameC].\\
\lbrack nameE] has [num] [relation] [object] than [nameD].\\
\\
Question：How many [object] does [nameE] have?\\\\

heurictic distractor：\\
\lbrack nameF] has [num] [relation] [object] than [nameA].\\
\lbrack nameG] has [num] [relation] [object] than [nameB].\\
\lbrack nameH] has [num] [relation] [object] than [nameC].\\
\\
distractor:\\
\lbrack nameI] has [num] [relation] [object] than [nameD].\\
\lbrack nameJ] has [num] [relation] [object] than [nameF].\\
\lbrack nameK] has [num] [relation] [object] than [nameF].\\
\lbrack nameL] has [num] [relation] [object] than [nameG].\\
\lbrack nameM] has [num] [relation] [object] than [nameG].\\
\lbrack nameN] has [num] [relation] [object] than [nameJ].\\
\lbrack nameO] has [num] [relation] [object] than [nameJ].\\
\lbrack nameP] has [num] [relation] [object] than [nameK].\\
\lbrack nameQ] has [num] [relation] [object] than [nameK].\\
\bottomrule
    \end{tabular}
        \caption{
    Template of artificial data in~\cref{subsec:exp2}.
    }
    \label{table:appendix_FL}
\end{table}
\endgroup
We prepare a template similar to a Table~\ref{table:appendix_FL} and assign values to the template according to the following steps:
\begin{itemize}
\item We create template as shown in  table\ref{table:appendix_FL1}.
\item Within the template, placeholders \lbrack nameA\rbrack \, to \lbrack nameQ\rbrack \, is filled randomly with names such as ``Alice'', ``Bob'', ``Carol'', ``Dave'', ``Eve'', ``Frank'', ``Grace'', ``Heidi'', ``Ivan'', ``Judy'', ``Kevin'', ``Larry'', ``Mallory'', ``Nancy'', ``Olivia'', ``Peggy'', ``Quentin'', ``Rob'', ``Sybil'', ``Trent'', ``Ursula'', "Victor'', ``Walter'', ``Xavier'', ``Yvonne'', ``Zoe''.
\item The placeholder \lbrack num\rbrack \, is filled with a random number from 0 to 100.
\item The placeholder \lbrack object\rbrack \, is filled randomly with items such as ``apples'', ``bananas'', ``grapes'', ``pencils'', ``books''.
\item The placeholder \lbrack relation\rbrack \, is assigned either ``more'' or ``fewer''.
\item Sentences within the context are shuffled randomly.
\item A distractor is inserted at a random position.
\end{itemize}

Then, using the following procedures, We create each expanded dataset. 
Each targeted heuristic strongly influences the heuristic distractors designed in this study. 
Each dataset consists of 300 problems.
\begin{itemize}
\item For the Overlap dataset, the values ``'s mother'', ``'s father'', ``'s son'', and ``'s neighborhood'' are appended to \lbrack nameE\rbrack \, and assigned respectively to \lbrack nameF\rbrack, \lbrack nameG\rbrack, and \lbrack nameH\rbrack. 
Each of \lbrack nameF\rbrack, \lbrack nameG\rbrack, and \lbrack nameH\rbrack \, hold different values.
\item For the Position dataset, the sentences with \lbrack nameF\rbrack, \lbrack nameG\rbrack, and \lbrack nameH\rbrack \, as the subjects have distractors inserted closer to the beginning of the context than the sentences with \lbrack nameB\rbrack, \lbrack nameC\rbrack, and \lbrack nameD\rbrack \, as the subjects. For other datasets, heuristic distractors are inserted at random positions.
\item For the Negative dataset, the form of the heuristic distractor is changed to a negative form.
\end{itemize}
In~\cref{subsec:exp2}, the method to identify which premises are used for reasoning was similar to that in~\cref{app:exp1_data}, relying on regular expressions.

\section{Mearuing accuracy in~\cref{subsec:exp1}}
This paper was mainly concerned with the frequency of distractor selection. To ensure that the model is not producing crappy output in these experiments, we measure the accuracy. Table~\ref{table:appendix_GSM8K_accuracy} below shows the accuracy rates of each model on GSM8K. 
Additionally, Table~\ref{table:appendix_FL_superficial_accuracy} below shows the accuracy rates of each model on artificial data.

From the Table~\ref{table:appendix_GSM8K_accuracy},~\ref{table:appendix_FL_superficial_accuracy}, it is shown that GPT-4 had the highest accuracy rates across the datasets, while Llama2 had the lowest. 
It is expected that these outcomes are due to differences in the number of parameters in the model.
\begin{table}[t]
\centering

\footnotesize
\begin{tabular}{lcccc}
\toprule
     & Base   & Over. & Pos.  & Neg. \\
\midrule
PaLM2   & 64.5\% & 59.2\%   & 60.5\%     & 71.1\%   \\
Llama2 & 30.3\% & 34.2\% & 30.3\% & 27.6\%\\
GPT-3.5 & 81.6\% & 64.5\% &81.6\%  & 82.9\%   \\
GPT-4   & 85.5\%  & 84.2\%   & 82.9\%  & 92.1\%   \\
\bottomrule
\end{tabular}
\caption{
The accuracy while solving GSM8K.
}
\label{table:appendix_GSM8K_accuracy}
\end{table}

\begin{table}[t]
\centering

\footnotesize
\begin{tabular}{lcccc}
\toprule
     & Base   & Over. & Pos. & Neg. \\
\midrule
PaLM2   & 60.0\% & 58.7\%   &77.0\%   &80.7\%   \\
Llama2 & 21.3\% & 14.3\% & 22.3\% & 20.0\%\\
GPT-3.5 & 84.6\% & 87.3\%   & 87.6\%     & 82.9\%   \\
GPT-4   & 98.7\%  & 94.0\%  & 98.0\%    &  99.7\%   \\
\bottomrule
\end{tabular}
\caption{
The accuracy while solving artificial reasoning tasks.
}
\label{table:appendix_FL_superficial_accuracy}
\end{table}

\begin{table}[t]
\centering

\footnotesize
\begin{tabular}{lccc}
\toprule
     & Over.$\uparrow$  & Pos.$\uparrow$  & Neg.$\downarrow$  \\
\midrule
PaLM2   & 41.0\%  & 14.0\%     & 5.7\%   \\
Llama2 & 82.0\%   & 93.3\%     & 28.0\%   \\
GPT-3.5 & 11.7\%   & 35.0\%     & 0.0\%  \\
GPT-4   & 0.0\%   & 0.0\%      & 0.0\%   \\
\bottomrule
\end{tabular}
\caption{
The frequency at which the model selected a distractor (i,e., $\Tilde{p}$) while solving artificial reasoning tasks after changing few-shot examples.
}
\label{table:appendix_FL_superficial_prompt}
\end{table}

\section{Generation settings}
When using GPT-3.5 and GPT-4, the settings are adjusted to temperature=0.0, frequency\_penalty=0, and presence\_penalty=0. Similarly, for PaLM2 and Llama2, the temperature is set to 0, with no sampling.

We use NVIDIA RTX A6000 (48GB) GPUs for inference with Llama2.

\section{Few-shot examples}
\label{app:prompt}
The few-shot examples for models regarding datasets GSM8K and artificial data are shown in the respective Tables~\ref{table:gsm8k_prompt},~\ref{table:fl_cot_prompt}.
\begingroup
\begin{table*}[t]
    \centering
    \small
    \begin{tabular}{p{0.9\linewidth}}
        \toprule
        Answer the context question using the following example.\\
        \vspace{1mm}
        \textbf{Context:} Leo's assignment was divided into three parts. Weng earns \$12 an hour for babysitting. It took Leo twice as long to finish the second part. Yesterday, she just did 50 minutes of babysitting.\\
        \textbf{Question:} How much did Weng earn?\\
        \textbf{Answer:} \\
        Weng earns 12/60 = 0.2 per minute.\\
        Working 50 minutes, she earned 0.2 x 50 = 10.\\
        The final answer is 10.\\
        \vspace{1mm}
        \textbf{Context:} Betty is saving money for a new wallet, which costs \$100. Betty has only half of the money she needs. Alice is saving money for a new wallet, which costs \$2000. Betty's parents decided to give Betty \$15 for that purpose, and her grandparents twice as much as her parents.
Question: How much more money does Betty need to buy the wallet?\\
        \textbf{Question:} How much more money does Betty need to buy the wallet?\\
        \textbf{Answer:} \\
        In the beginning, Betty has only 100 / 2 = 50.\\
        Betty's parents gave her 15.\\
        Betty's grandparents gave her 15 * 2 = 30.\\
        This means Betty needs 100 - 50 - 15 - 30 = 5 more.\\
        The final answer is 5.\\
        \vspace{1mm}
        \textbf{Context:} Julie is reading a 120-page book. Yesterday, Julie was able to read 12 pages, and today, she read twice as many pages as yesterday. Julie's mother makes \$18.00 an hour.\\
        \textbf{Question:} If Julie wants to read half of the remaining pages tomorrow, how many are left to read?\\
        \textbf{Answer:}\\
        Julie read 12 x 2 = <<12*2=24>>24 pages today\\
        So, she was able to read a total of 12 + 24 = 36 pages since yesterday.\\
        There are 120 - 36 = 84 pages left to be read.\\
        Since she wants to read half of the remaining pages tomorrow, she should read 84/2 = 42 pages.\\
        The final answer is 42.\\
        \vspace{1mm}
        \textbf{Context:} James writes a 2-page letter to 4 different friends who live in America twice a week. James writes a 3-page letter to 2 different friends who live in Japan twice a week.\\
        \textbf{Question:} How many pages does James write each friend who lived in Japan for a year?\\
        \textbf{Answer:}\\
        He writes each friend 3*2=6 pages a week.\\
        So, he writes 6*2=12 pages every week.\\
        That means he writes 12*52=624 pages a year.\\
        The final answer is 624.\\
        \bottomrule
    \end{tabular}
        \caption{
    Examples of input given when solving GSM8K.
    }
    \label{table:gsm8k_prompt}
\end{table*}
\endgroup
\begingroup
\begin{table*}[t]
    \centering
    \small
    \begin{tabular}{p{0.9\linewidth}}
        \toprule
        Answer the context question using the following example.\\
        \vspace{1mm}
        \textbf{Context:} Walter has -22 apples. Ursula has 3 more apples than Walter. Victor has 3 more apples than Ursula. Quentin has 2 more apples than Ursula. Nancy has 3 more apples than Walter. Zoe has 3 more apples than Nancy. Heidi has 3 more apples than Nancy. Carol's mother has 4 apples. Xavier has 3 more apples than Carol's mother. Peggy has 4 more apples than Xavier. Dave has 13 more apples than Xavier. Bob has 1 more apples than Carol's mother. Alice has 3 more apples than Bob. Sybil has 56 more apples than Bob.\\
        \textbf{Question:} How many apples does Dave have?\\
        \textbf{Answer:} \\
        Carol's mother has 4 apples, and Xavier has 3 more apples than Carol's mother. So, Xavier has 4+3=7 apples.\\
        Xavier has 7 apples, and Dave has 13 more apples than Xavier. So, Dave has 7+13=20 apples. 
The final answer is 20.\\
        \vspace{1mm}
        \textbf{Context:} Alice has 92 more bananas than Mallory. Victor has 10 fewer bananas than Walter. Xavier has 59 more bananas than Sybil. Yvonne has 79 more bananas than Sybil. Judy has 23 more bananas than Alice. Dave has 60 more bananas than Victor. Quentin has 35 fewer bananas than Peggy. Heidi has 95 more bananas than Victor. Ursula doesn't have 32 more bananas than Peggy. Larry has 17 fewer bananas than Alice. Zoe has 58 fewer bananas than Yvonne. Ivan has 43 fewer bananas than Yvonne. Walter has 43 fewer bananas than Mallory. Nancy has 34 bananas. Grace has 41 more bananas than Xavier. Mallory has 55 fewer bananas than Nancy. Sybil has 3 fewer bananas than Nancy. Peggy has 50 more bananas than Walter. Trent has 33 fewer bananas than Xavier.\\
        \textbf{Question:} How many bananas does Quentin have?\\
        \textbf{Answer:} \\
        Nancy has 34 bananas, and Mallory has 55 fewer bananas than Nancy. So, Mallory has 34-55=-21 bananas.\\
        Mallory has -21 bananas, and Walter has 43 fewer bananas than Mallory. So, Walter has -21-43=-64 bananas.\\
        Walter has -64 bananas, and Peggy has 50 more bananas than Walter. So, Peggy has -64+50=-14 bananas.\\
        Peggy has -14 bananas, and Quentin has 35 fewer bananas than Peggy. So, Quentin has -14-35=-49 bananas.\\
        The final answer is -49.\\
        \vspace{1mm}
        \textbf{Context:} Zoe has 10 more apples than Yvonne's son. Eve has 2 apples. Yvonne's son has 3 more apples than Eve. Quentin has 3 more apples than Yvonne. Yvonne has 3 fewer apples than Zoe. Alice has 3 more apples than Grace. Trent has 34 more apples than Zoe. Ivan has 3 apples. Ursula has 3 more apples than Zoe. Grace has 3 apples. Xavier doesn't have 3 more apples than Ivan.\\
        \textbf{Question:} How many apples does Yvonne have?\\
        \textbf{Answer:}\\
        Eve has 2 apples, and Yvonne's son has 3 more apples than Eve. So, Yvonne's son has 2+3=5 apples.\\
        Yvonne's son has 5 apples, and Zoe has 10 more apples than Yvonne's son. So, Zoe has 5+10=15 apples.\\
        Zoe has 15 apples, and Yvonne has 3 fewer apples than Zoe. So, Yvonne has 15-3=12 apples.\\
        The final answer is 12.\\
        \vspace{1mm}
        \textbf{Context:} Kevin's friend has 33 fewer grapes than Rob. Ivan has 43 more grapes than Victor. Victor has 33 fewer grapes than Kevin's friend. Ursula has 75 fewer grapes than Zoe. Alice has 11 more grapes than Eve. Dave has 11 more grapes than Eve. Olivia has 29 more grapes than Kevin's friend. Mallory has 97 more grapes than Olivia. Judy has 78 more grapes than Olivia. Rob has 55 grapes. Frank has 70 fewer grapes than Heidi. Eve has 84 fewer grapes than Sybil. Xavier has 36 more grapes than Heidi. Sybil has 55 fewer grapes than Trent. Kevin has 43 fewer grapes than Zoe. Heidi has 61 fewer grapes than Trent. Zoe has 88 more grapes than Sybil. Trent has 40 more grapes than Rob. Walter has 38 more grapes than Victor.\\
        \textbf{Question:} How many grapes does Kevin have?\\
        \textbf{Answer:}\\
        Rob has 55 grapes, and Trent has 40 more grapes than Rob. So, Trent has 55+40=95 grapes.\\
        Trent has 95 grapes, and Sybil has 55 fewer grapes than Trent. So, Sybil has 95-55=40 grapes.\\
        Sybil has 40 grapes, and Zoe has 88 more grapes than Sybil. So, Zoe has 40+88=128 grapes.\\
        Zoe has 128 grapes, and Kevin has 43 fewer grapes than Zoe. So, Kevin has 128-43=85 grapes.\\
        The final answer is 85.\\
        \bottomrule
    \end{tabular}
        \caption{
    Examples of input given when solving an artificial dataset.
    }
    \label{table:fl_cot_prompt}
\end{table*}
\endgroup

\section{Effect of few-shot examples}
We investigate whether the few-shot examples trigger the model's heuristic. Specifically, we replace the few-shot examples in the following ways to study the relationship between the model's heuristic and its inputs:
\\
1. We change the few-shot examples to induce Overlap (as shown in Table~\ref{table:fl_overlap_prompt}) and examine whether this increases the reasoning frequency with the use of distractors in the Overlap dataset compared to what is shown in Table~\ref{table:FL_all_result}.
\\
2. We change the few-shot examples to induce Position (as shown in Table~\ref{table:fl_positional_prompt}) and check if there's an increase in reasoning frequency with the use of distractors in the Position dataset compared to Table~\ref{table:FL_all_result}.
\\
3. We change the few-shot examples to induce Negative (as shown in Table~\ref{table:fl_lexical_prompt}), and investigate if there's a decrease in reasoning frequency with the use of distractors in the Negative dataset compared to Table~\ref{table:FL_all_result}.

We measure the frequency of selecting $\Tilde{p}$ in~\cref{subsec:exp1}. 
The results are presented in Table~\ref{table:appendix_FL_superficial_prompt}. 
As shown in Tables~\ref{table:FL_all_result},~\ref{table:appendix_FL_superficial_prompt}, although the few-shot examples fed into the models such as GPT-3.5, GPT-4, and PaLM2 was changed, there was no significant change in reasoning frequency as described. 
This suggests that the model’s heuristic does not merely mimic the examples provided as input.
On the other hand, the Llama2 model was more prone to being misled by changes in input, and smaller models demonstrated a reduced capacity to reach the correct answers.

\section{Effects of increasing the number of distructors}
We investigate whether the same results could be obtained when the number of distractors that do not include heuristics increases. Specifically, we prepare the same settings as in~\cref{subsec:exp2} and add distractors that do not include heuristics. We define the distractors that include heuristics as $\tilde{p}_{t,\text{heuristic}}$ and those that do not as $\tilde{p}_{t,\text{non-heuristic}}$. The ratio of $\tilde{p}_{t,\text{heuristic}}$ to $\tilde{p}_{t,\text{non-heuristic}}$ in the problem text is 1:8. In this experiment, we use a model that could produce reasonable output even when the context length increases. Specifically, we use only the GPT-3.5-turbo and GPT-4.

\begin{figure}[t]
\centering
\includegraphics[width=\linewidth]{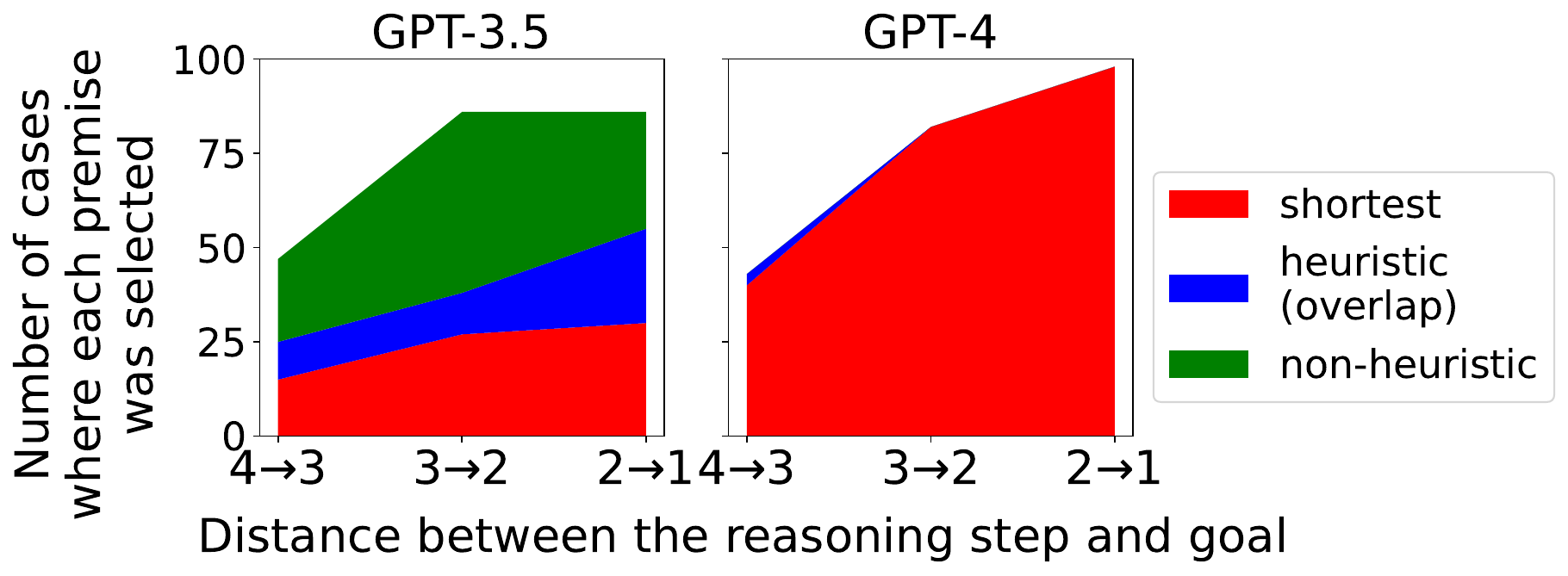}
\caption{Number of cases where a particular distractor is selected (y-axis: $r$) in each reasoning step (x-axis: $d$). The heuristic to be included in the distructor is overlap.}
\label{fig:appendix_overlap}
\end{figure}
\begin{figure}[t]
\centering
\includegraphics[width=\linewidth]{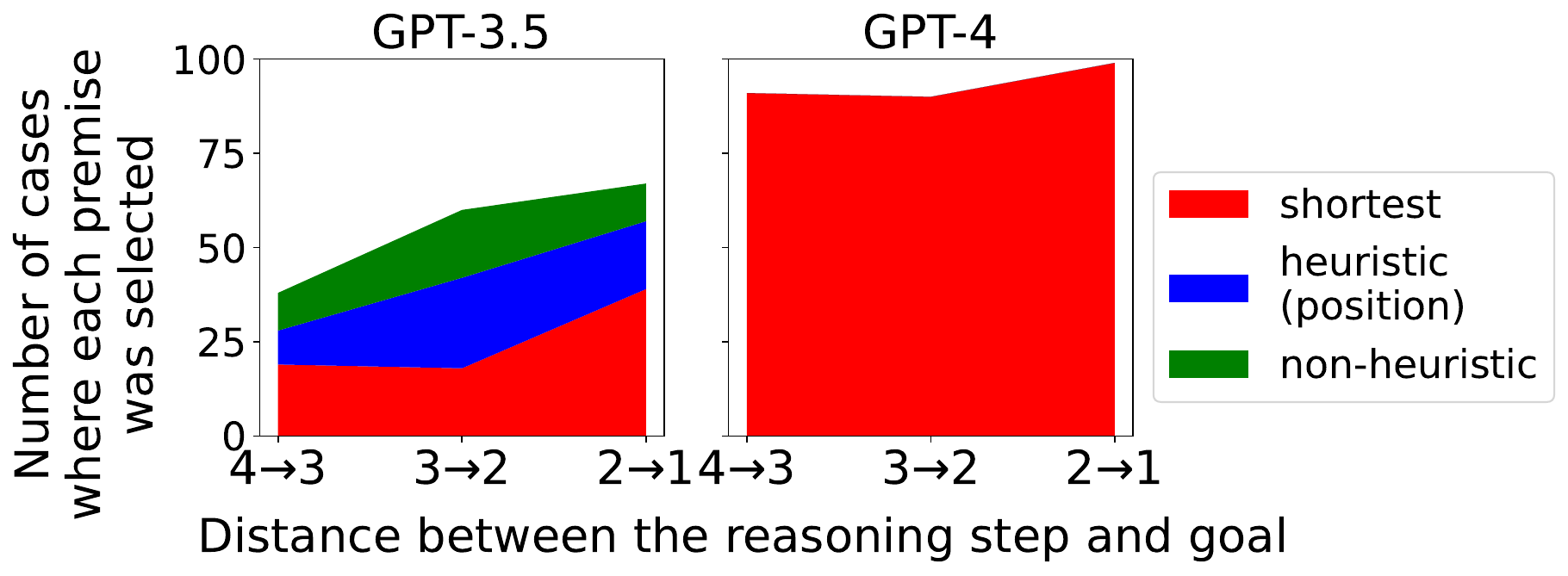}
\caption{Number of cases where a particular distractor is selected (y-axis: $r$) in each reasoning step (x-axis: $d$). The heuristic to be included in the distructor is position.}
\label{fig:appendix_position}
\end{figure}

Figures~\cref{fig:appendix_overlap,fig:appendix_position} show how many times out of 100 times the model selected $\tilde{p}_{t,\text{heuristic}}$, $\tilde{p}_{t,\text{non-heuristic}}$, and $h^*_t$ at each step. Figures~\cref{fig:appendix_overlap,fig:appendix_position} show the experimental results for the cases where the heuristics included in the distructor are overlap and position, respectively.
Figures~\cref{fig:appendix_overlap,fig:appendix_position} show that the number of cases in which the shortest path is selected increases as the goal is reached.
The ratio of distructors that include heuristics to those that do not is 1:8, but the ratio of distructors that include heuristics when distructors are selected is higher than $1/(1+8)$. Therefore, this suggests that premises are being selected using heuristics.
Except for the GPT-3.5 (overlap) results, the number of cases where a destructor is selected decreases as the goal is approached.
These results are as with the results for~\ref{subsec:exp2}. The results of GPT-3.5 (overlap) may indicate that the increase in the number of distructors has reduced the focus on distructors, including heuristics.

\begingroup
\begin{table*}[t]
    \centering
    \small
    \begin{tabular}{p{0.9\linewidth}}
        \toprule
        Answer the context question using the following example.\\
        \vspace{1mm}
        \textbf{Context:} Context: Walter has -22 apples. Ursula has 3 more apples than Walter. Victor has 3 more apples than Ursula. Quentin has 2 more apples than Ursula. Nancy has 3 more apples than Walter. Zoe has 3 more apples than Nancy. Heidi has 3 more apples than Nancy. Dave's mother has 4 apples. Dave's father has 3 more apples than Dave's mother. Peggy has 4 more apples than Dave's father. Dave has 13 more apples than Dave's father. Bob has 1 more apples than Carol's mother. Alice has 3 more apples than Bob. Sybil has 56 more apples than Bob.\\
        \textbf{Question:} How many apples does Dave have?\\
        \textbf{Answer:} \\
        Dave's mother has 4 apples, and Dave's father has 3 more apples than Dave's mother. So, Dave's father has 4+3=7 apples.\\
        Dave's father has 7 apples, and Dave has 13 more apples than Dave's father. So, Dave has 7+13=20 apples.\\
        The final answer is 10.\\
        \vspace{1mm}
        \textbf{Context:} Alice has 92 more bananas than Quentin's mother. Victor has 10 fewer bananas than Walter. Xavier has 59 more bananas than Sybil. Yvonne has 79 more bananas than Sybil. Judy has 23 more bananas than Alice. Dave has 60 more bananas than Victor. Quentin has 35 fewer bananas than Quentin's father. Heidi has 95 more bananas than Victor. Ursula doesn't have 32 more bananas than Quentin's father. Larry has 17 fewer bananas than Alice. Zoe has 58 fewer bananas than Yvonne. Ivan has 43 fewer bananas than Yvonne. Walter has 43 fewer bananas than Quentin's mother. Nancy has 34 bananas. Grace has 41 more bananas than Xavier. Quentin's mother has 55 fewer bananas than Nancy. Sybil has 3 fewer bananas than Nancy. Quentin's father has 50 more bananas than Walter. Trent has 33 fewer bananas than Xavier.\\
        \textbf{Question:} How many bananas does Quentin have?\\
        \textbf{Answer:} \\
        Nancy has 34 bananas, and Quentin's mother has 55 fewer bananas than Nancy. So, Quentin's mother has 34-55=-21 bananas.\\
        Quentin's mother has -21 bananas, and Walter has 43 fewer bananas than Quentin's mother. So, Walter has -21-43=-64 bananas.\\
        Walter has -64 bananas, and Quentin's father has 50 more bananas than Walter. So, Quentin's father has -64+50=-14 bananas.\\
        Quentin's father has -14 bananas, and Quentin has 35 fewer bananas than Quentin's father. So, Quentin has -14-35=-49 bananas.\\
        The final answer is -49.\\
        \vspace{1mm}
        \textbf{Context:} Yvonne's father has 10 more apples than Yvonne's son. Eve has 2 apples. Yvonne's son has 3 more apples than Eve. Quentin has 3 more apples than Yvonne. Yvonne has 3 fewer apples than Yvonne's father. Alice has 3 more apples than Grace. Trent has 34 more apples than Yvonne's father. Ivan has 3 apples. Ursula has 3 more apples than Yvonne's father. Grace has 3 apples. Xavier has 3 more apples than Ivan.\\
        \textbf{Question:} How many apples does Yvonne have?\\
        \textbf{Answer:}\\
        Eve has 2 apples, and Yvonne's son has 3 more apples than Eve. So, Yvonne's son has 2+3=5 apples.\\
        Yvonne's son has 5 apples, and Yvonne's father has 10 more apples than Yvonne's son. So, Yvonne's father has 5+10=15 apples.\\
        Yvonne's father has 15 apples, and Yvonne has 3 fewer apples than Yvonne's father. So, Yvonne has 15-3=12 apples.\\
        The final answer is 12.\\
        \vspace{1mm}
        \textbf{Context:} Kevin's friend has 33 fewer grapes than Rob. Ivan has 43 more grapes than Victor. Victor has 33 fewer grapes than Kevin's friend. Ursula has 75 fewer grapes than Zoe. Alice has 11 more grapes than Eve. Dave has 11 more grapes than Eve. Olivia has 29 more grapes than Kevin's friend. Mallory has 97 more grapes than Olivia. Judy has 78 more grapes than Olivia. Rob has 55 grapes. Frank has 70 fewer grapes than Heidi. Eve has 84 fewer grapes than Kevin's neighborhood. Xavier has 36 more grapes than Heidi. Kevin's neighborhood has 55 fewer grapes than Kevin's friend. Kevin has 43 fewer grapes than Kevin's mother. Heidi has 61 fewer grapes than Kevin's friend. Kevin's mother has 88 more grapes than Kevin's neighborhood. Kevin's friend has 40 more grapes than Rob. Walter has 38 more grapes than Victor.\\
        \textbf{Question:} How many grapes does Kevin have?\\
        \textbf{Answer:}\\
        Rob has 55 grapes, and Kevin's friend has 40 more grapes than Rob. So, Kevin's friend has 55+40=95 grapes.\\
        Kevin's friend has 95 grapes, and Kevin's neighborhood has 55 fewer grapes than Kevin's friend. So, Kevin's neighborhood has 95-55=40 grapes.\\
        Kevin's neighborhood has 40 grapes, and Kevin's mother has 88 more grapes than Kevin's neighborhood. So, Kevin's mother has 40+88=128 grapes.\\
        Kevin's mother has 128 grapes, and Kevin has 43 fewer grapes than Kevin's mother. So, Kevin has 128-43=85 grapes.\\
        The final answer is 85.\\
        \bottomrule
    \end{tabular}
        \caption{
    Examples of input given when solving the Overlap dataset.
    }
    \label{table:fl_overlap_prompt}
\end{table*}
\endgroup
\begingroup
\begin{table*}[t]
    \centering
    \small
    \begin{tabular}{p{0.9\linewidth}}
        \toprule
        Answer the context question using the following example.\\
        \vspace{1mm}
        \textbf{Context:} Carol's mother has 4 apples. Xavier has 3 more apples than Carol's mother. Dave has 13 more apples than Xavier. Walter has -22 apples. Ursula has 3 more apples than Walter. Victor has 3 more apples than Ursula. Quentin has 2 more apples than Ursula. Nancy has 3 more apples than Walter. Zoe has 3 more apples than Nancy. Heidi has 3 more apples than Nancy. Peggy has 4 more apples than Xavier. Bob has 1 more apples than Carol's mother. Alice has 3 more apples than Bob. Sybil has 56 more apples than Bob.\\
        \textbf{Question:} How many apples does Dave have?\\
        \textbf{Answer:} \\
        Carol's mother has 4 apples, and Xavier has 3 more apples than Carol's mother. So, Xavier has 4+3=7 apples.\\
        Xavier has 7 apples, and Dave has 13 more apples than Xavier. So, Dave has 7+13=20 apples. 
The final answer is 20.\\
        \vspace{1mm}
        \textbf{Context:} Nancy has 34 bananas. Mallory has 55 fewer bananas than Nancy. Walter has 43 fewer bananas than Mallory. Peggy has 50 more bananas than Walter. Quentin has 35 fewer bananas than Peggy. Alice has 92 more bananas than Mallory. Victor has 10 fewer bananas than Walter. Xavier has 59 more bananas than Sybil. Yvonne has 79 more bananas than Sybil. Judy has 23 more bananas than Alice. Dave has 60 more bananas than Victor. Heidi has 95 more bananas than Victor. Ursula doesn't have 32 more bananas than Peggy. Larry has 17 fewer bananas than Alice. Zoe has 58 fewer bananas than Yvonne. Ivan has 43 fewer bananas than Yvonne. Grace has 41 more bananas than Xavier. Sybil has 3 fewer bananas than Nancy. Trent has 33 fewer bananas than Xavier.\\
        \textbf{Question:} How many bananas does Quentin have?\\
        \textbf{Answer:} \\
        Nancy has 34 bananas, and Mallory has 55 fewer bananas than Nancy. So, Mallory has 34-55=-21 bananas.\\
        Mallory has -21 bananas, and Walter has 43 fewer bananas than Mallory. So, Walter has -21-43=-64 bananas.\\
        Walter has -64 bananas, and Peggy has 50 more bananas than Walter. So, Peggy has -64+50=-14 bananas.\\
        Peggy has -14 bananas, and Quentin has 35 fewer bananas than Peggy. So, Quentin has -14-35=-49 bananas.\\
        The final answer is -49.\\
        \vspace{1mm}
        \textbf{Context:} Eve has 2 apples. Yvonne's son has 3 more apples than Eve. Zoe has 10 more apples than Yvonne's son. Yvonne has 3 fewer apples than Zoe. Alice has 3 more apples than Grace. Quentin has 3 more apples than Yvonne. Trent has 34 more apples than Zoe. Ivan has 3 apples. Ursula has 3 more apples than Zoe. Grace has 3 apples. Xavier has 3 more apples than Ivan.\\
        \textbf{Question:} How many apples does Yvonne have?\\
        \textbf{Answer:}\\
        Eve has 2 apples, and Yvonne's son has 3 more apples than Eve. So, Yvonne's son has 2+3=5 apples.\\
        Yvonne's son has 5 apples, and Zoe has 10 more apples than Yvonne's son. So, Zoe has 5+10=15 apples.\\
        Zoe has 15 apples, and Yvonne has 3 fewer apples than Zoe. So, Yvonne has 15-3=12 apples.\\
        The final answer is 12.\\
        \vspace{1mm}
        \textbf{Context:} Rob has 55 grapes. Trent has 40 more grapes than Rob. Sybil has 55 fewer grapes than Trent. Zoe has 88 more grapes than Sybil. Kevin has 43 fewer grapes than Zoe. Kevin's friend has 33 fewer grapes than Rob. Ivan has 43 more grapes than Victor. Victor has 33 fewer grapes than Kevin's friend. Ursula has 75 fewer grapes than Zoe. Alice has 11 more grapes than Eve. Dave has 11 more grapes than Eve. Olivia has 29 more grapes than Kevin's friend. Mallory has 97 more grapes than Olivia. Judy has 78 more grapes than Olivia. Frank has 70 fewer grapes than Heidi. Eve has 84 fewer grapes than Sybil. Xavier has 36 more grapes than Heidi. Heidi has 61 fewer grapes than Trent. Walter has 38 more grapes than Victor.\\
        \textbf{Question:} How many grapes does Kevin have?\\
        \textbf{Answer:}\\
        Rob has 55 grapes, and Trent has 40 more grapes than Rob. So, Trent has 55+40=95 grapes.\\
        Trent has 95 grapes, and Sybil has 55 fewer grapes than Trent. So, Sybil has 95-55=40 grapes.\\
        Sybil has 40 grapes, and Zoe has 88 more grapes than Sybil. So, Zoe has 40+88=128 grapes.\\
        Zoe has 128 grapes, and Kevin has 43 fewer grapes than Zoe. So, Kevin has 128-43=85 grapes.\\
        The final answer is 85.\\
        \bottomrule
    \end{tabular}
        \caption{
    Examples of input given when solving the Position dataset.
    }
    \label{table:fl_positional_prompt}
\end{table*}
\endgroup
\begingroup
\begin{table*}[t]
    \centering
    \small
    \begin{tabular}{p{0.9\linewidth}}
        \toprule
        Answer the context question using the following example.\\
        \vspace{1mm}
        \textbf{Context:} Walter doesn't have -22 apples. Ursula has 3 more apples than Walter. Victor has 3 more apples than Ursula. Quentin has 2 more apples than Ursula. Nancy doesn't have 3 more apples than Walter. Zoe has 3 more apples than Nancy. Heidi doesn't have 3 more apples than Nancy. Carol's mother has 4 apples. Xavier has 3 more apples than Carol's mother. Peggy has 4 more apples than Xavier. Dave has 13 more apples than Xavier. Bob doesn't have 1 more apples than Carol's mother. Alice has 3 more apples than Bob. Sybil has 56 more apples than Bob.\\
        \textbf{Question:} How many apples does Dave have?\\
        \textbf{Answer:} \\
        Carol's mother has 4 apples, and Xavier has 3 more apples than Carol's mother. So, Xavier has 4+3=7 apples.\\
        Xavier has 7 apples, and Dave has 13 more apples than Xavier. So, Dave has 7+13=20 apples. 
The final answer is 20.\\
        \vspace{1mm}
        \textbf{Context:} Alice has 92 more bananas than Mallory. Victor has 10 fewer bananas than Walter. Xavier has 59 more bananas than Sybil. Yvonne doesn't have 79 more bananas than Sybil. Judy doesn't have 23 more bananas than Alice. Dave has 60 more bananas than Victor. Quentin has 35 fewer bananas than Peggy. Heidi has 95 more bananas than Victor. Ursula doesn't have 32 more bananas than Peggy. Larry doesn't have 17 fewer bananas than Alice. Zoe has 58 fewer bananas than Yvonne. Ivan has 43 fewer bananas than Yvonne. Walter has 43 fewer bananas than Mallory. Nancy has 34 bananas. Grace doesn't have 41 more bananas than Xavier. Mallory has 55 fewer bananas than Nancy. Sybil doesn't have 3 fewer bananas than Nancy. Peggy has 50 more bananas than Walter. Trent doesn't have 33 fewer bananas than Xavier.\\
        \textbf{Question:} How many bananas does Quentin have?\\
        \textbf{Answer:} \\
        Nancy has 34 bananas, and Mallory has 55 fewer bananas than Nancy. So, Mallory has 34-55=-21 bananas.\\
        Mallory has -21 bananas, and Walter has 43 fewer bananas than Mallory. So, Walter has -21-43=-64 bananas.\\
        Walter has -64 bananas, and Peggy has 50 more bananas than Walter. So, Peggy has -64+50=-14 bananas.\\
        Peggy has -14 bananas, and Quentin has 35 fewer bananas than Peggy. So, Quentin has -14-35=-49 bananas.\\
        The final answer is -49.\\
        \vspace{1mm}
        \textbf{Context:} Zoe has 10 more apples than Yvonne's son. Eve has 2 apples. Yvonne's son has 3 more apples than Eve. Quentin has 3 more apples than Yvonne. Yvonne has 3 fewer apples than Zoe. Alice has 3 more apples than Grace. Trent has 34 more apples than Zoe. Ivan has 3 apples. Ursula has 3 more apples than Zoe. Grace has 3 apples. Xavier doesn't have 3 more apples than Ivan.\\
        \textbf{Question:} How many apples does Yvonne have?\\
        \textbf{Answer:}\\
        Eve has 2 apples, and Yvonne's son has 3 more apples than Eve. So, Yvonne's son has 2+3=5 apples.\\
        Yvonne's son has 5 apples, and Zoe has 10 more apples than Yvonne's son. So, Zoe has 5+10=15 apples.\\
        Zoe has 15 apples, and Yvonne has 3 fewer apples than Zoe. So, Yvonne has 15-3=12 apples.\\
        The final answer is 12.\\
        \vspace{1mm}
        \textbf{Context:} Kevin's friend has 33 fewer grapes than Rob. Ivan doesn't have 43 more grapes than Victor. Victor doesn't have 33 fewer grapes than Kevin's friend. Ursula has 75 fewer grapes than Zoe. Alice has 11 more grapes than Eve. Dave has 11 more grapes than Eve. Olivia doesn't have 29 more grapes than Kevin's friend. Mallory has 97 more grapes than Olivia. Judy doesn't have 78 more grapes than Olivia. Rob has 55 grapes. Frank has 70 fewer grapes than Heidi. Eve has 84 fewer grapes than Sybil. Xavier doesn't have 36 more grapes than Heidi. Sybil has 55 fewer grapes than Trent. Kevin has 43 fewer grapes than Zoe. Heidi has 61 fewer grapes than Trent. Zoe has 88 more grapes than Sybil. Trent has 40 more grapes than Rob. Walter has 38 more grapes than Victor.\\
        \textbf{Question:} How many grapes does Kevin have?\\
        \textbf{Answer:}\\
        Rob has 55 grapes, and Trent has 40 more grapes than Rob. So, Trent has 55+40=95 grapes.\\
        Trent has 95 grapes, and Sybil has 55 fewer grapes than Trent. So, Sybil has 95-55=40 grapes.\\
        Sybil has 40 grapes, and Zoe has 88 more grapes than Sybil. So, Zoe has 40+88=128 grapes.\\
        Zoe has 128 grapes, and Kevin has 43 fewer grapes than Zoe. So, Kevin has 128-43=85 grapes.\\
        The final answer is 85.\\
        \bottomrule
    \end{tabular}
        \caption{
    Examples of input given when solving the Neg dataset.
    }
    \label{table:fl_lexical_prompt}
\end{table*}
\endgroup

\section{Usage of Writing Assistance}
We use publicly available writing assistance tools, including Grammarly, to refine the language for readability.
\end{document}